\documentclass[preprint, 12pt]{elsarticle}

\usepackage[utf8]{inputenc} 
\usepackage[T1]{fontenc}    
\usepackage{hyperref}       
\usepackage{url}            
\usepackage{booktabs}       
\usepackage{amsfonts}       
\usepackage{nicefrac}       
\usepackage{microtype}      
\usepackage{lipsum}		
\usepackage{graphicx}
\usepackage{subfigure}
\usepackage{natbib}
\usepackage{doi}
\usepackage{amsmath}
\usepackage{bm}
\usepackage{algorithm}
\usepackage{algpseudocode}
\usepackage{geometry}
\usepackage{array}
\usepackage{booktabs}
\usepackage{multirow}
\usepackage{xcolor}
\usepackage{makecell}
\usepackage{amsmath,amsfonts}
\newcommand{\argmin}{\operatornamewithlimits{argmin}}
\newcommand{\bx}{\bm{x}}

\newcommand{\by}{\bm{y}}

\newcommand{\bD}{\textbf{D}}

\newcommand{\bL}{\textbf{L}}
\newcommand{\bC}{\textbf{C}}

\newcommand{\comment}[1]{}

\comment{
\author{ \href{https://orcid.org/0000-0000-0000-0000}{\includegraphics[scale=0.06]{orcid.pdf}\hspace{1mm}Yilin Zhuang}\\
	Department of Aerospace Engineering\\
	University of Michigan\\
	Ann Arbor, MI 48105 \\
	\texttt{ylzhuang@umich.edu} \\
	\AND
	{\includegraphics[scale=0.06]{orcid.pdf}\hspace{1mm}Sibo Cheng}\\
	CEREA\\
	\'{E}cole des Ponts and EDF R\&D, Institut Polytechnique de Paris\\
    \^Ile-de-France	  \\
	\texttt{sibo.cheng@enpc.fr} \\
	  \And
    {\includegraphics[scale=0.06]{orcid.pdf}\hspace{1mm}Karthik Duraisamy}\\
	Department of Aerospace Engineering\\
	University of Michigan\\
	Ann Arbor, MI 48105 \\
	\texttt{kdur@umich.edu} \\
}

}

\hypersetup{
pdftitle={A template for the arxiv style},
pdfsubject={q-bio.NC, q-bio.QM},
pdfauthor={David S.~Hippocampus, Elias D.~Striatum},
pdfkeywords={First keyword, Second keyword, More},
}

\begin{document}

\begin{frontmatter}

\title{Spatially-Aware Diffusion Models with Cross-Attention for Global Field Reconstruction with Sparse Observations}

\author[label1]{Yilin Zhuang}
\author[label2]{Sibo Cheng}
\author[label1]{Karthik Duraisamy\corref{cor1}}

\cortext[cor1]{Corresponding author.}
\ead{kdur@umich.edu}
\affiliation[label1]{
organization={Department of Aerospace Engineering, University of Michigan},
city={Ann Arbor},
postcode={48105},
state={Michigan},
country={United States}
}
\affiliation[label2]{
organization={CEREA, \'{E}cole des Ponts and EDF R\&D, Institut Polytechnique de Paris},
city={\^Ile-de-France},
country={France}
}

\begin{abstract}
Diffusion models have gained attention for their ability to represent complex distributions and incorporate uncertainty, making them ideal for robust predictions in the presence of noisy or incomplete data. In this study, we develop and enhance score-based diffusion models in field reconstruction tasks, where the goal is to estimate complete spatial fields from partial observations. We introduce a condition encoding approach to construct a tractable mapping between observed and unobserved regions using a learnable integration of sparse observations and interpolated fields as an inductive bias. With refined sensing representations and an unraveled temporal dimension, our method can handle arbitrary moving sensors and effectively reconstruct fields. Furthermore, we conduct a comprehensive benchmark of our approach against a deterministic interpolation-based method across various static and time-dependent PDEs. Our study attempts to addresses the gap in strong baselines for evaluating performance across varying sampling hyperparameters, noise levels, and conditioning methods. Our results show that diffusion models with cross-attention and the proposed conditional encoding generally outperform other methods under noisy conditions, although the deterministic method excels with noiseless data. Additionally, both the diffusion models and the deterministic method surpass the numerical approach in accuracy and computational cost for the steady problem. We also demonstrate the ability of the model to capture possible reconstructions and improve the accuracy of fused results in covariance-based correction tasks using ensemble sampling. 
\end{abstract}

\comment{
\begin{highlights}
    \item We propose a cross-attention diffusion model for global field reconstruction.
    \item We integrate Voronoi-tessellated fields and sensing positions as an inductive bias.
    \item Method can handle arbitrary moving sensors and
effectively reconstruct spatio-temporal fields.
    \item Cross-attention diffusion model outperforms others in noisy observation setting. 
    \item Ensemble mean in probabilistic reconstructions converges to deterministic output.
\end{highlights}
}



\end{frontmatter}

\section{Introduction} \label{sec:intro}
 The global field reconstruction problem, which involves reconstructing a full field from partial observations, is underdetermined and has long been a challenge across various science and engineering domains~\cite{fukami2021global, shen2015missing, 8281621}. 
Various numerical and deep-learning methods have been proposed to address this challenge, including Kriging~\cite{kleijnen2009kriging}, iterative Kalman filtering-based methods~\cite{huang2022iterated,bocquet2014iterative}, Voronoi-tessellation Convolutional Neural Networks (VCNNs)~\cite{fukami2021global}, and Physics-Informed Neural Networks~\cite{raissi2019physics}. 

Among classical numerical methods for solving the field reconstruction tasks, Gaussian process~\cite{rasmussen2003gaussian} and its variants, such as the ensemble Kalman filter~\cite{evensen1994sequential, bocquet2011ensemble,iglesias2013ensemble} and extended Kalman filter~\cite{jazwinski2007stochastic, huang2022iterated}, are commonly used statistical methods for approximating fields using Gaussian kernels. For optimization-based approaches, they are often combined with model reduction techniques to manage high-dimensional fields and perform field reconstruction~\cite{huang2022iterated, angell2018inferring, gu2017application, gherlone2012shape}. However, these methods remain computationally expensive, and the optimization formulation can become intractable for time-dependent PDEs.

Various deep learning frameworks have been developed for field reconstruction tasks. The VCNN~\cite{fukami2021global, zhang2024uncertainty} is a convolutional neural network that uses Voronoi tessellation to map interpolated fields to reconstructed fields. Voronoi tessellation describes a class of interpolation methods that maps point data to a field. 
Neural operators function by simultaneously learning differential operators and field solutions. Variants such as Physics-Informed Neural Operators (PINOs)~\cite{li2024physics} and Latent Neural Operators (LNOs)~\cite{wang2024latent} are also capable of solving inverse problems. Another commonly used method is the Physics-Informed Neural Network (PINN)~\cite{raissi2019physics}, which leverages automatic differentiation to solve PDEs. Several variants of PINNs and automatic differentiation-based methods have been proposed to address inverse problems~\cite{zhu2019physics, smith2022hyposvi, santos2023development}. These methods are typically deterministic, and uncertainty quantification is often performed by injecting noise into the observations. For a fixed set of observations, fields reconstructed by deterministic methods are fixed and do not support uncertainty quantification.

Generative models, derived from probabilistic learning and variational inference, have emerged as a powerful class of methods for generating new samples from data distribution. In the context of field reconstruction, generative models map an initial distribution, typically Gaussian, to the target data distribution~\cite{liu2022flow}, conditioned on the observed fields. Previous work has demonstrated that Generative Adversarial Networks (GANs) can reconstruct patches of turbulence data based on observations of the remaining fields~\cite{buzzicotti2021reconstruction}. However, it has also been shown that diffusion models can outperform GANs in image synthesis and are easier to train~\cite{dhariwal2021diffusion, yang2023diffusion}. Additionally, diffusion models have demonstrated exceptional ability in learning complex data distributions across diverse domains~\cite{hoDenoisingDiffusionProbabilistic2020, liu2024sora, li2022diffusionlm}, making them an ideal candidate for performing probabilistic generation. 

Several works have applied diffusion models to solve forward~\cite{jacobsen2023cocogen, huang2024diffusionpde} and inverse problems~\cite{jacobsen2023cocogen, bastek2024physicsinformed, huang2024diffusionpde, dasgupta2024conditional}, as well as the incorporation of physical residual to enhance the accuracy of generated fields~\cite{jacobsen2023cocogen, chung2022diffusion, huang2024diffusionpde}. Most of these works are based on full-field diffusion models, which are capable of directly backpropagating the physical loss and integrating seamlessly with partial observations when solving inverse problems. However, there has also been a growing interest in applying latent diffusion models to physics field generation tasks~\cite{gao2024generative, du2024confildconditionalneuralfield}.

There are various ways to perform field reconstruction tasks with diffusion models that condition on observations. In the image processing domain, guided sampling or inpainting is frequently used because these techniques can be directly applied to trained diffusion models~\cite{song2023pseudoinverse, mardani2023variational, chung2022diffusion}. Guided sampling works by using the diffusion model to reconstruct unknown regions in the field. When applied in the physical domain, guided sampling is often combined with physical information to achieve physically realistic results~\cite{jacobsen2023cocogen, huang2024diffusionpde, shu2023physics, dasgupta2024conditional}. Some studies have also adopted the classifier-free guidance (CFG)~\cite{hoClassifierFreeDiffusionGuidance2022} method to incorporate sensing information~\cite{haitsiukevich2024diffusion, huang2024diffda} or physical information~\cite{du2024confildconditionalneuralfield, jacobsen2023cocogen} as guiding information into diffusion models. This guiding information is typically integrated by augmenting the noise scale embedding and the latent representations of the fields. The cross-attention method is another frequently used approach for conditioning in the image processing domain~\cite{radford2021learning, rout2024beyond}. Cross-attention is a variant of self-attention~\cite{vaswani2017attention} where the attention mechanism is applied between the image latent and the conditioning embedding. Compared to the augmentation performed in CFG, cross-attention has been shown to handle complex conditioning information~\cite{po2024state}, which could help capture variations in observation positions. Santos et al.~\citep{santos2023development} have investigated applying cross-attention-based deterministic method for field reconstruction tasks, and demonstrated promising results. However, to the best of our knowledge, cross-attention has not yet been explored in diffusion models for field reconstruction tasks.

Despite the success of diffusion models in previous studies, comprehensive benchmarking of their performance in field reconstruction tasks remains limited, specifically in terms of comparisons against a strong baseline. Furthermore, previous research has not thoroughly explored the comparisons between different conditioning methods when applying diffusion models to physical fields. In our earlier work~\cite{jacobsen2023cocogen}, we enforced physical consistency in the generated fields during the reverse sampling process. However,  reverse sampling trajectories could be disrupted if the scales of the coefficients for the physical and sensing residuals are not properly managed. This issue is particularly problematic for time-dependent PDEs, where it is difficult to precisely evaluate the physical residual due to mismatches between the time intervals of saved snapshots and the actual time steps.

In this study, we propose a conditional encoding approach that leverages inductive bias and observation positions to construct a tractable mapping between observed and unobserved regions in a full-field diffusion model for field reconstruction tasks. We conduct an extensive benchmark comparing the diffusion model with the interpolation-based deterministic model, VCNN~\cite{fukami2021global}. Furthermore, we evaluate different conditioning methods, including guided sampling (or inpainting), CFG~\cite{hoClassifierFreeDiffusionGuidance2022}, and cross-attention~\cite{chen2021crossvit}, while analyzing the effects of sampling hyperparameters. Our results indicate that applying cross-attention in conjunction with our proposed condition encoding block results in superior performance compared to the other two conditioning methods.

Field reconstruction from sparse sensor data is closely related to super-resolution tasks, where the goal is to recover high-resolution fields from limited observations, as commonly seen in fluid dynamics applications~\cite{fukami2023super}. Recent studies have applied diffusion models to super-resolution problems, demonstrating their ability to handle complex nonlinear structures~\cite{du2024confildconditionalneuralfield, shu2023physics}. Although our work focuses on field reconstruction, the proposed cross-attention method can also be applied to super-resolution tasks, given the similarity in input data structure and the need for high-fidelity reconstruction from sparse observations.

The implemented diffusion models are based on a U-Net~\cite{ronneberger2015u} architecture, which includes additional connections between down-sampling and up-sampling blocks enhancing its capability compared to standard CNNs. To ensure a fair comparison, we adapt the VCNN to VT-UNet and perform self-attention in the middle block of the UNet. Our benchmark includes one static and three time-dependent PDEs: the Darcy flow, shallow water equation, diffusion-reaction equation, and compressible Navier-Stokes equations. 

The deep learning models are also compared with the numerical iterative Kalman filtering method on the Darcy flow problem. 
Additionally, we demonstrate the diffusion model's capability to estimate ensemble mean and uncertainty, which can be incorporated into a numerical covariance inverse model~\cite{tandeo2020review}. This capability is demonstrated on the shallow water equations~\cite{cheng2024efficient}, where we show that the fused result can be improved using the uncertainty estimated by the diffusion model. The code for our models and experiments is publicly available in the Git repository: \href{https://github.com/tonyzyl/DiffusionReconstruct}{https://github.com/tonyzyl/DiffusionReconstruct}.


The rest of this paper is organized as follows. In Section~\ref{sec:methods}, we review the problem formulation and the underlying architecture of the diffusion model with the condition encoding block. The benchmark results of diffusion models with various hyperparameters are compared with the deterministic method in Section~\ref{sec:res}. Finally, we conclude with highlights and discuss potential future improvements in Section~\ref{sec:conclusion}.

\section{Methods} \label{sec:methods}

\subsection{Problem formulation}

Consider a two dimensional squared domain $\Omega \in \mathbb{R}^{N_d \times N_d}$, where $N_d$ is the grid size. Let $\bm{x} \in \mathbb{R}^{N_c \times N_d \times N_d}$ denote the fields on $\Omega$, where $N_c$ is the number of fields. We denote $\mathcal{M} \in \mathbb{R}^{N_c \times N_d \times N_d}$ as the observation matrix with the one-hot encoding:
\begin{equation}
    \mathcal{M}_{i,j,k} = \begin{cases}
        1 & \text{if } (j,k) \in \text{Observed points} \\
        0 & \text{otherwise}
    \end{cases}
\end{equation}
Here, we assume that the observed points across all fields have the same coordinates. We also define the unobserved matrix $\mathcal{M}^c$ such that $\bm{x} = (\mathcal{M} \odot \bm{x}) \oplus (\mathcal{M}^c \odot \bm{x})$, where $\odot$ and $\oplus$ denote the element-wise multiplication and addition, respectively. Let $\mathcal{H} \colon \mathbb{R}^{N_c \times N_d \times N_d} \to \mathbb{R}^{N_c \times N_{obs}}$ denote the observation operator, and we have the observed data as $\bm{y} = \mathcal{H}(\bm{x})$.

Let $\{s_{c,1}, s_{c,2}, \ldots, s_{c,N_{obs}}\} \subseteq \Omega $ denote the set of Voronoi-tessellated field  for the variable $\bm{x}_c \in \mathbb{R}^{N_d \times N_d}$. Each sub-region $s_{c,i}$ is defined as:
\begin{equation}
\begin{split}
    s_{c,i} &= \{x \in \Omega \mid \| x - \text{Pos}(\bm{y}_{c,i}) \| \leq \| x - \text{Pos}(\bm{y}_{c,j}) \|, \forall j \neq i\}, \\
    &\text{with } s_{c,i}(x) = \bm{y}_{c,i}, \forall x \in s_{c,i}.
\end{split}
\end{equation}
where Pos denotes the position of the observed point for field $\bm{x}_c$. Let $\bm{q}$ denote the reconstructed field, and the reconstructions using VT-UNet, unconditional diffusion and conditional diffusion models can be obtained as: $\bm{q} = F_{\text{VT}}(\{s_{c,i}\})$, $\bm{q} = F_{\text{Diff}}(\bm{\epsilon}, \bm{y})$, and $\bm{q} = F_{\text{CondDiff}}(\bm{\epsilon}, \{s_{c,i}\}, \mathcal{M}\odot \bm{x})$, respectively. Here, we slightly abuse the notation for diffusion models because $\bm{q}$ is generated through iterative calls to the diffusion model, and \(\bm{\epsilon}\) denotes the randomized field initialization. The VT-UNet model is trained to minimize the mean squared error, $\mathbb{E}_{\bm{x}, \bm{y}} \left[ \| \bm{q} - \bm{x} \|_2^2 \right]$

\subsection{Diffusion model with spatial feature cross attention}

The forward map of diffusion models is a tractable transformation where noise is gradually added, and the reverse map is approximated by neural networks to generate the reconstructed fields~\cite{esser2024scaling}. We denote the data distribution as \(\pi_0\) and the random noise as \(\pi_1 \sim \mathcal{N}(0, \bm{I})\). Let \(\bm{x}_0\) be the initial data sample. Its intermediate representations \(\bm{x}_t\) at timesteps \(t \in [0,1]\) can be obtained through the following transformation:
\begin{equation}
    \bm{x}_t = a_t\bm{x}_0 + b_t \bm{\epsilon}, \quad \text{where } \bm{\epsilon} \sim \mathcal{N}(0, \bm{I}),
    \label{eq:edm_traj}
\end{equation}
where \(a_t\) and \(b_t\) are the parameters of the transformation. Here, the timestep $t$ is an artificial notation for describing the mapping between the data distribution and the Gaussian prior, rather than physical time. Various choices exist for these transformation parameters~\cite{song2020denoising, hoDenoisingDiffusionProbabilistic2020, song2021scorebased, liu2022flow}. The Elucidating Diffusion Model (EDM) framework~\cite{karras2022elucidating} can be regarded as a special case of variance-exploding (VE) formulation~\cite{kingma2024understanding} and it can be expressed as:
\begin{equation}
    \bm{x} = \bm{x}_0 + \sigma_t \bm{\epsilon},
\end{equation}
where \(\sigma_t\) denotes the noise level, sampled from a log-normal distribution during training.  For simplicity, we will drop the subscript of $\bm{x}_t$ and $\sigma_t$. One advantage of the VE formulation is its capability to handle unevenly distributed data, which is common in physical fields. Even after normalizing with the mean and standard deviation of the training dataset, physical fields can exhibit significant variability, with some regions being highly positive and others highly negative, despite having a mean close to zero. The variance-exploding formulation is well-suited to address this issue, as it can accommodate large noise scales.

We also tested a diffusion model with noise prediction using the variance-preserving (VP)~\cite{song2021scorebased} formulation on the Darcy flow problem. We found the model trained with VP formulation struggled to generate the unevenly distributed fields. One possible reason is the sampled Gaussian noise typically has a smaller magnitude than the variability in the uneven regions. In this case, the noise level may not be large enough to capture the variability in the data distribution, leading to poor generation performance starting from the Gaussian prior.

For the reverse sampling process, instead of solving the stochastic differential equation (SDE), Song et al.~\cite{song2021scorebased} proposed solving the following probability flow (PF) ordinary differential equation (ODE):
\begin{equation}
    d\bm{x} = \left[ \bm{f}(\bm{x}, t) - \frac{1}{2}g(t)^2\nabla \log p_t(\bm{x}; t) \right] dt,
\end{equation}
where \(\bm{f}\) and \(g\) are the drift and diffusion functions, respectively. \(\log p_t(\bm{x}; t)\) is the score function, which is the gradient of the log-likelihood of the data distribution at time \(t\) with respect to the data sample \(\bm{x}\)~\cite{hyvarinen2005estimation}. For generating physical fields, the PF ODE is preferred over the SDE due to its deterministic nature, which ensures a more tractable generation process~\cite{jacobsen2023cocogen}.

Let \(D(\bm{x};\sigma)\) denote the denoiser function that is optimized by the following training objective~\cite{karras2022elucidating} to minimize the \(L_2\) denoising error:
\begin{equation}\label{eq:score_match}
    \mathbb{E}_{x_0 \sim p_{\text{data}}} \mathbb{E}_{\bm{n} \sim \mathcal{N} (0,\sigma^2 I)} \|D(\bm{x}_0 + \bm{n}; \sigma) - \bm{x}_0\|^2_2, \quad \text{with}\, \nabla_{\bm{x}} \log p_t(\bm{x}; \sigma) = \frac{D(\bm{x}; \sigma) - \bm{x}}{\sigma^2}
\end{equation}
where \(\bm{n}\) denotes the added noise. Instead of approximating the denoiser function directly with neural network, it has been shown that scaling the output of denoising estimator with respect to the noise level, $\sigma$, improves overall performance. The following scaling scheme is utilized in the loss function~\cite{karras2022elucidating}:
\begin{equation}
    D_\theta(\bm{x};\sigma)=c_{skip}(\sigma)\bm{x}+c_{out}(\sigma)F_\theta(c_{in}\left(\sigma)\bm{x};c_{noise}(\sigma)\right)
\end{equation}
\begin{equation}
   \mathbb{E}_{\sigma, \bm{x}_0,\bm{n}} \left[\lambda(\sigma)c_{out}(\sigma)^2 \|F_\theta \left( c_{in}(\sigma)\cdot (\bm{x}_0+\bm{n});c_{noise}(\sigma)\right)-\frac{1}{c_{out}(\sigma)} \left( \bm{x}_0-c_{skip}(\sigma)\cdot(\bm{x}_0+\bm{n})\right) \|_2^2 \right]
   \label{eq:edm_loss}
\end{equation}
where \(\lambda(\sigma)\) is a positive weighting function, \(c_{out}(\sigma)\), \(c_{noise}(\sigma)\), and \(c_{in}(\sigma)\) are scaling factors. The function $F_\theta$ represents the neural network parameterized by $\theta$. To generate the full-field solution, we solve the following deterministic ODE, derived by substituting \(\sigma(t)=t\) as the noise schedule in Equation~\eqref{eq:score_match}
\begin{equation}\label{eq:pfode}
    d\bm{x}_{-} = -t\nabla_{\bm{x}} \log p_t(\bm{x};\sigma)dt= \frac{\bm{x}-D_\theta(\bm{x};\sigma)}{t}dt
\end{equation}
We utilize the multi-step and predictor-corrector methods to solve the ODE.

With access to partial measurements of the field, Song et al.~\cite{song2021scorebased} proved that the score function can be approximated as:
\begin{equation} \label{eq:score_partial_obs}
    \nabla_{\bm{z}} \log p_t(\bm{z}_t|\bm{y})\approx \nabla_{\bm{z}} \log p_t(\bm{z}_t|\mathcal{M}^c \odot \hat{\bm{x}}_t)=\nabla_{\bm{z}} \log p_t\left([\bm{z}_t\oplus(\mathcal{M}^c \odot \hat{\bm{x}}_t)\right)
\end{equation}
where $\bm{z}_t=\mathcal{M}^c \odot \bm{x}_t$ defines a new diffusion process of the unknown fields, and $\mathcal{M}^c \odot \hat{\bm{x}}$ denotes a random sample from $p_t(\mathcal{M}^c \odot \bm{x}_t|\bm{y})$.

\begin{figure}[!htb]
    \centering
    \includegraphics[width=\textwidth]{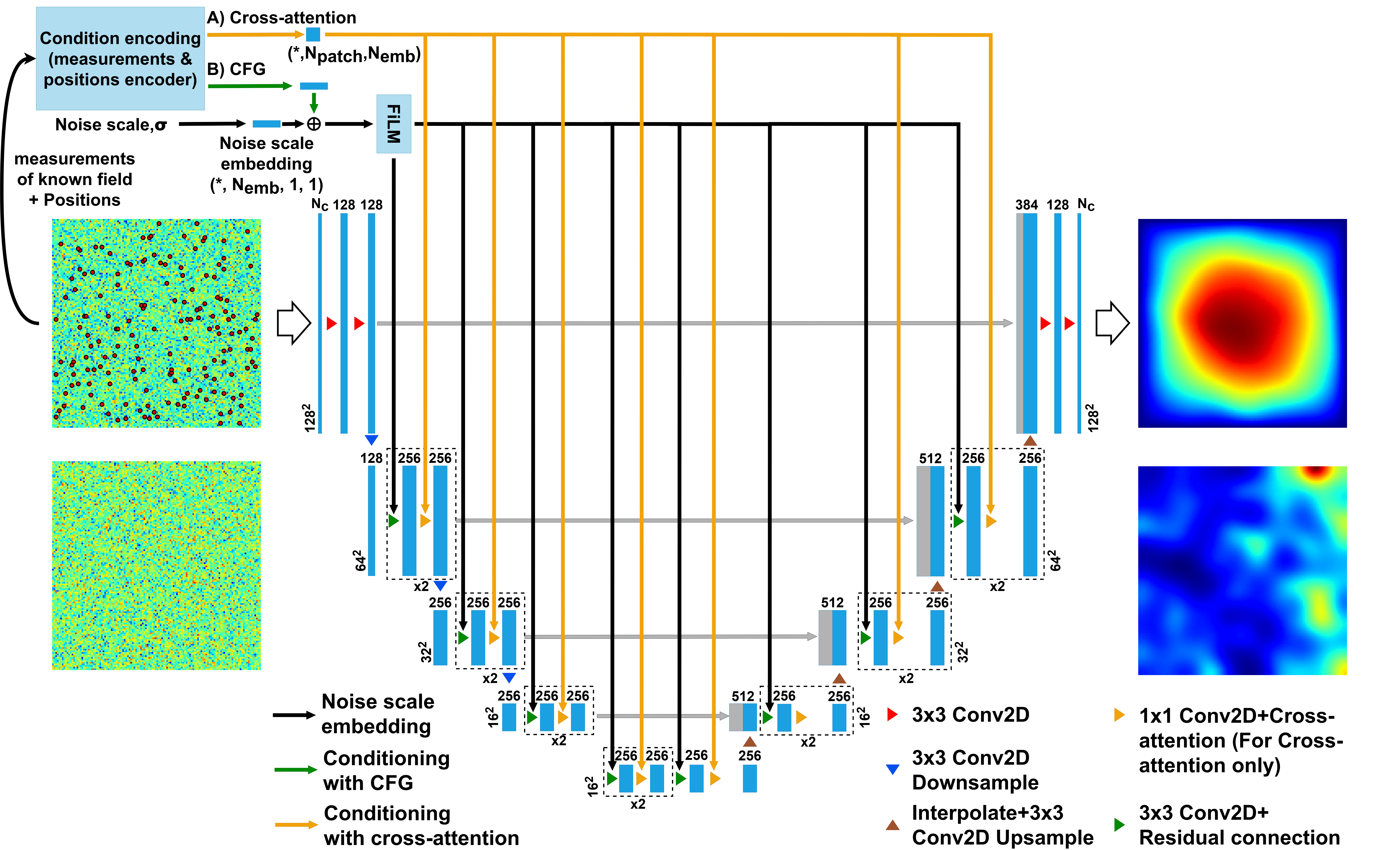}
    \caption{Schematic of the proposed condition encoding block with the UNet-based diffusion model, $F_\theta$, and two ways of encoding sensor information: (A) cross-attention and (B) classifier-free guidance.}
    \label{fig:scheme}
\end{figure}

Using partial observations as conditioning information, we tested three conditioning methods: guided sampling, classifier-free guidance, and cross-attention. A schematic of the latter two methods, along with the proposed condition encoding block, is shown in Figure~\ref{fig:scheme}. The guided sampling method is based on the inpainting approach, where the full fields are initially filled with noise, and an unconditional model is trained to denoise the fields. For the guided reverse sampling process, the unobserved field is updated by Equation~\eqref{eq:pfode}, $\mathcal{M}^c \odot d\bm{z}_{-}$, and the observed field is updated by~\cite{lugmayr2022repaint}:
\begin{equation}
    \mathcal{M} \odot \bm{x}_{t-1} = \mathcal{M} \odot \bm{x}_{0} + \sigma_{t-1}\bm{\epsilon}
\end{equation}

For CFG~\cite{hoClassifierFreeDiffusionGuidance2022}, the pooled embedding of the conditioning information is combined with the noise scale embedding  using the Feature-wise Linear Modulation (FiLM)~\cite{perez2017film} to generate the denoised fields. FiLM performs learnable modulations on the hidden state using the conditional information, offering an effective and flexible way of modulating the hidden state. In the cross-attention approach~\cite{chen2021crossvit}, cross-attention is applied between the embedding of the conditioning information and the hidden states, $h$, of the diffusion model. Let $E_\phi$ denote the condition encoding block. The cross-attention has the same formulation as self-attention but with different matrix assignments:
\begin{equation}
\text{Attention}(Q, K, V) = \text{softmax}\left(\frac{QK^T}{\sqrt{d_k}}\right) V,
\end{equation}
where \( Q = W_q h \), \( K = W_k E_\phi(x) \), and \( V = W_v E_\phi(x) \). Here, the query (\( Q \)) is derived from the hidden states of the diffusion model, while the key (\( K \)) and value (\( V \)) are derived from the condition encoding block \( E_\phi(x) \). \( W_q \), \( W_k \), and \( W_v \) are learned projection matrices, and \( d_k \) is the dimensionality of the key.

Mathematically, conditioning via cross-attention can be regarded as a form of CFG, the new score function of the unobserved field can be expressed as:
\begin{equation}
\begin{split}
    \nabla_{\bm{z}} \log p_t(\bm{z}_t|E_\phi(\bm{y}))= & \nabla_{\bm{z}} \log p_t(\bm{z}_t|E_\phi(\bm{y}_{\text{null}})) + \\
    & \gamma \cdot \left( \nabla_{\bm{z}} \log p_t(\bm{z}_t|E_\phi(\bm{y})) - \nabla_{\bm{z}} \log p_t(\bm{z}_t|E_\phi(\bm{y}_{\text{null}})) \right)
    \label{eq:cfg}
\end{split}
\end{equation}

where $\gamma$ is the guidance scale, set to 1, and $E_\phi(\bm{y}_{\text{null}})$ denotes the unconditional encoded state. Compared to Equation~\eqref{eq:score_partial_obs}, we rely on the condition encoding block to capture the encoded representation and establish a tractable mapping between observed and unobserved regions. We set the CFG and cross-attention diffusion models to capture $p(\bm{z}(t)|E_\phi(\bm{y}))$, while retaining Equation~\eqref{eq:edm_loss} as the training objective for the encoder block to extract the observations. During the reverse sampling process, we update only the unobserved regions of the field.

\begin{figure}[!htb]
    \centering
    \includegraphics[width=\textwidth]{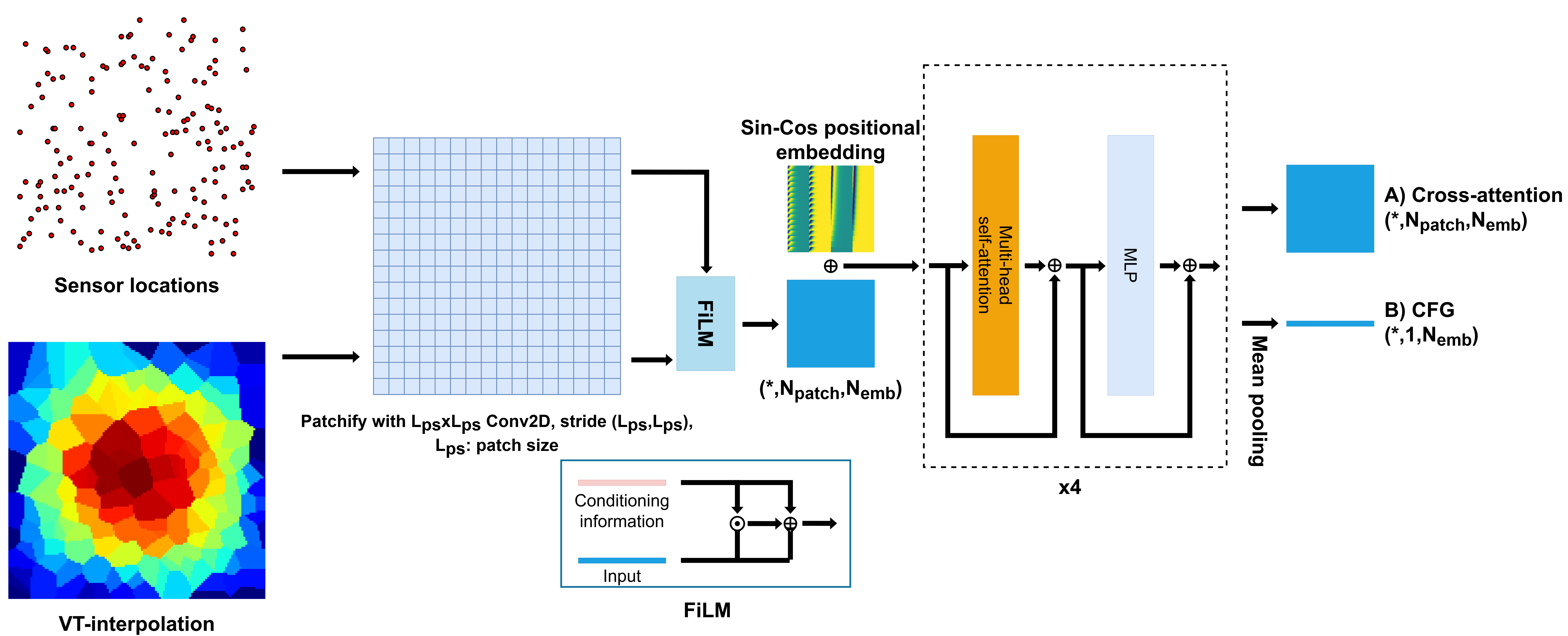}
    \caption{Schematic of the proposed condition encoding block. For CFG, mean-pooling is performed to reduce the dimensionality and to combine it with the noise scale embedding.}
    \label{fig:block_scheme}
\end{figure}

The proposed condition encoding block processes information from the Voronoi-tessellated fields and sensing positions, integrating their patched embeddings using FiLM. The Voronoi-tessellated fields serve as an inductive bias and have previously been applied to diffusion model for super-resolution tasks~\cite{shu2023physics}. The encoded states are further refined through a multilayer perceptron (MLP) and self-attention layers. A schematic of the proposed encoding block is shown in Figure~\ref{fig:block_scheme}. The adapted VT-UNet architecture, which mirrors that of the diffusion model, maps the Voronoi-tessellated fields to the reconstructed fields. For time-dependent PDEs, the temporal dimension is unraveled during training, with no physical time provided as conditioning information. This unravelling approach aligns with the use of Voronoi tessellation for field inversion~\cite{fukami2021global} and effectively handles field reconstruction from moving sensors. 

Each model is trained for 100,000 steps on 8 Nvidia H100 GPUs, with weights updated using an Exponential Moving Average (EMA). We do not include a validation step for saving the best weights. Additional details on the training and implementation are provided in~\ref{app:train}. A discussion of how the trained models can be finetuned to handle different resolutions is provided in~\ref{app:res}.

\subsection{Data assimilation as posterior fine-tuning}
\label{sec:DA}

The prediction of the physical field can be further enhanced using data assimilation (DA) algorithms based on Bayesian methods. Let $\bm{x}_{b,\Tilde{t}}$ denote the predicted control vector (also known as the background state) and $\bm{y}_{\Tilde{t}}$ denote the sparse observation at time $\Tilde{t}$~\cite{Carrassi2017}. In this section, $\Tilde{t}$ denotes the physical time in simulations. 
Variational DA aims to find the optimal compromise between $\bm{x}_{b,\Tilde{t}}$ and $\bm{y}_{\Tilde{t}}$ by minimizing the cost function $J_{\Tilde{t}}$, defined as:

\begin{align}
    J_{\Tilde{t}}(\bm{x})&=\frac{1}{2}(\bm{x}-\bm{x}_{b,\Tilde{t}})^T\textbf{B}_{\Tilde{t}}^{-1}(\bm{x}-\bm{x}_{b,\Tilde{t}}) + \frac{1}{2}(\bm{y}_{\Tilde{t}}-\mathcal{H}(\bm{x}_{\Tilde{t}})).^T \textbf{R}_{\Tilde{t}}^{-1} (\bm{y}_{\Tilde{t}}-\mathcal{H}(\bm{x}_{\Tilde{t}})) \label{eq_3dvar}\\
   &=\frac{1}{2}\vert \vert\bm{x}-\bm{x}_{b,\Tilde{t}}\vert \vert^2_{\textbf{B}_{\Tilde{t}}^{-1}}+\frac{1}{2}\vert \vert\bm{y}_{\Tilde{t}}-\mathcal{H}(\bm{x})\vert \vert^2_{\textbf{R}_{\Tilde{t}}^{-1}} \notag
\end{align}
where the operator $(\cdot)^T$ in Equation~\eqref{eq_3dvar} indicates the transpose. The error covariance matrices associated with $\bm{x}_{b,t}$ and $\bm{y}_{\Tilde{t}}$ are denoted by $\textbf{B}_{\Tilde{t}}$ and $\textbf{R}_{\Tilde{t}}$, respectively:

\begin{align}
     \textbf{B}_{\Tilde{t}} = \textrm{Cov}(\bm{x}_{b,\Tilde{t}} - \bm{x}_{\textrm{true},\Tilde{t}}, \bm{x}_{b,\Tilde{t}} - \bm{x}_{\textrm{true},\Tilde{t}}), \quad
     \textbf{R}_{\Tilde{t}} = \textrm{Cov}(\mathcal{H}(\bm{x}_{\textrm{true},\Tilde{t}})-\bm{y}_{\Tilde{t}}, \mathcal{H}(\bm{x}_{\textrm{true},\Tilde{t}})-\bm{y}_{\Tilde{t}}),
 \end{align}
where $\bm{x}_{\textrm{true},\Tilde{t}}$ represents the ground truth. 
Equation~\eqref{eq_3dvar} represents the three-dimensional variational (3D-Var) approach. The analysis state $\bm{x}_{a,\Tilde{t}}$ corresponds to the point at which the cost function in Equation~\eqref{eq_3dvar} reaches its minimum, that is,
  \begin{align}
    \bm{x}_{a,\Tilde{t}} = \underset{\bm{x}}{\argmin} \Big(J_{\Tilde{t}}(\bm{x})\Big) \label{eq:argmin}.
 \end{align}

Typically, DA assumes that the background error (i.e., prior estimation error) and the observation error are uncorrelated. Since the diffusion model prediction is generated using the observation points, we adopt the DA framework here only as a posterior fine-tuning tool. In this context, the background error covariance $\textbf{B}_{\Tilde{t}}$ can be empirically estimated from the ensemble output of the diffusion model, a task that is challenging for deterministic machine learning approaches~\cite{cheng2023machine}. Therefore, the approach proposed in this paper also addresses the bottleneck of prior and posterior error estimation in inverse modeling~\cite{tandeo2020review}.
To improve efficiency and effectively capture the spatial correlation of physical fields, DA is conducted within the reduced-order space of Principal Component Analysis (PCA). The details of this reduced order DA algorithm are provided in~\ref{sec:app_DA}. 

\subsection{Benchmark Problems}

We benchmark the performance of the diffusion model with different conditioning methods against the adapted VT-UNet on three fluid-like systems and one static system. Below, we provide a brief overview of the benchmark problem setups, with more detailed information on the data sources and generation procedures available in~\ref{app:data}. A summary of the benchmark problems is provided in Table~\ref{tab:benchmark}. The three time-dependent PDEs selected here cover advection and diffusion dynamics for fluid-like systems, as well as non-linear reaction dynamics. The static problem is chosen because it is a common benchmark for reconstructing fields from correlated observations.

\begin{table}[!htb]
\centering
\caption{Summary of datasets used for benchmarking the diffusion model with different conditioning methods.}
\label{tab:benchmark}
\begin{tabular}{c c c c c c}
\toprule
\textbf{PDE} & \makecell{\bm{$N_d$}} & \makecell{\bm{$N_t$}} & \makecell{\textbf{Boundary} \\ \textbf{Condition}} & \makecell{\textbf{Number of} \\ \textbf{Simulations}} & \makecell{\textbf{Data} \\ \textbf{Source}} \\ 
\midrule
Darcy flow & $128\times 128$ & N/A & Dirichlet & 10,000 &~\cite{huang2022iterated} \\ 
Shallow water & $64\times 64$ & 50 & Periodic & 250 &~\cite{cheng2024efficient} \\ 
2D Diffusion reaction & $128\times 128$ & 101 & Neumann & 1,000 &~\cite{takamoto2022pdebench} \\ 
2D Compressible Navier Stokes & $128\times 128$ & 21 & Periodic & 10,000 &~\cite{takamoto2022pdebench} \\ 
\bottomrule
\end{tabular}
\end{table}

\subsubsection{Darcy Flow}
The Darcy flow equations describe the relationship between fluid pressure, $p(\mathbf{x})$, and the permeability, $\alpha(\mathbf{x}, \theta)$, of a porous medium through which the fluid moves. The pressure and the permeability field are governed by the following relationships:

\begin{align}
    -\nabla \cdot (\alpha(\mathbf{x}, \theta)\nabla p(\mathbf{x})) & = f_s(\mathbf{x}), \;\;\; \mathbf{x} \in D, \\
    p(\mathbf{x}) & = 0, \;\;\; \mathbf{x} \in \partial D
\end{align}

The permeability field is generated using a Karhunen-Loève Expansion (KLE) of a Gaussian random field. The dataset is generated with 128 modes, and the corresponding pressure field is computed. In this problem, only partial observations of the pressure field are available. The numerical iterative Kalman filtering method~\cite{huang2022iterated} optimizes the coefficients of 64 modes to minimize the observation error. We use the code provided in~\citep{huang2022iterated} to generate 10,000 samples, with observation points evenly spaced across the pressure field. The boundary conditions are set to Dirichlet.

\subsubsection{Shallow water}

The shallow water equations describe a non-linear wave propagation problem defined over a spatial domain with three variables: water height, \( h(\mathbf{x}) \) (in \textit{mm}), x-velocity, \( \mathbf{u} \), and y-velocity, \( \mathbf{v} \). The equations are given by:
\begin{align}\label{eq:shallow}
    \frac{\partial h}{\partial t} + \nabla \cdot (h\mathbf{u}) & = 0,\\
    \frac{\partial \mathbf{u}}{\partial t} +  \frac{\partial h}{\partial x} + b\mathbf{u} & = 0,  \\
    \frac{\partial \mathbf{v}}{\partial t} +  \frac{\partial h}{\partial y} + b\mathbf{v} & = 0, \\
    \mathbf{u}_{t=0} & = 0, \\
    \mathbf{v}_{t=0} & = 0 
\end{align}
The simulations represent a dam break scenario, where a column of water is released at a random location within the domain. The boundary conditions are periodic, and We use the data simulated in~\citep{cheng2024efficient}, with partial observations of all three fields available.

\subsubsection{2D Diffusion-reaction}

The 2D diffusion-reaction system consists of two fields: the concentrations of an activator and an inhibitor. The equations for this system are given by:

\begin{align}
    \frac{\partial u}{\partial t} = & D_u \frac{\partial^2 u}{\partial x^2} + D_u \frac{\partial^2 u}{\partial y^2} + R_u, \\
    \frac{\partial v}{\partial t} = & D_v \frac{\partial^2 v}{\partial x^2} + D_v \frac{\partial^2 v}{\partial y^2} + R_v,
\end{align}
where \( u \) and \( v \) are the activator and inhibitor fields, respectively, with diffusion coefficients \( D_u =1\times 10^{-3} \) and \( D_v =5 \times 10^{-3} \). The reaction terms \( R_u \) and \( R_v \) are defined by the FitzHugh-Nagumo equations:
\begin{align}
    R_u (u, v) = & \, u - u^3 - k - v \\
    R_v (u, v) = & \, u - v
\end{align}
where \( k = 5 \times 10^{-3} \). The initial concentration at each point in both fields follows a Gaussian distribution. We use the data simulated in~\citep{takamoto2022pdebench}, with partial observations of both fields available. The boundary conditions are set to Neumann.

\subsubsection{2D Compressible Navier-Stokes (CFD)}

The compressible Navier-Stokes equations describe the motion of a compressible fluid. The equations for the 2D compressible Navier-Stokes system are given by:

\begin{align}
& \frac{\partial \rho}{\partial t} + \nabla \cdot (\rho \mathbf{v}) = 0, \\
& \rho \left(\frac{\partial \mathbf{v}}{\partial t} + \mathbf{v} \cdot \nabla \mathbf{v}\right) = -\nabla p + \eta \Delta \mathbf{v} + \left(\zeta + \frac{\eta}{3}\right)\nabla(\nabla \cdot \mathbf{v}), \\
& \frac{\partial}{\partial t} \left( \epsilon + \frac{\rho v^2}{2} \right) + \nabla \cdot \left[ \left(\epsilon + p + \frac{\rho v^2}{2}\right) \mathbf{v} - \mathbf{v} \cdot \sigma' \right] = 0,
\end{align}
where \( \rho \) is the density, \( \mathbf{v} \) is the velocity, \( p \) is the pressure, \( \sigma'\) is the viscous stress tensor, \( \eta \) and \( \zeta \) are the shear and bulk viscosities, respectively, and \( \epsilon \) is the internal energy. The initial conditions are constructed by a randomly initialized superposition of sinusoidal waves. We use the data simulated in~\citep{takamoto2022pdebench}, with partial observations of all four fields available. The selected dataset has \( \eta = \zeta = M = 0.1 \). The boundary conditions are set to periodic.

\section{Numerical Results} \label{sec:res}

For performing the field reconstruction tasks, we select the ratio of observed data points to be 0.3\% and 1.37\% for the Darcy flow problem, corresponding to 49 and 225 observed points on a $128\times128$ grid, respectively. To align with the original numerical approach, these observed points are evenly spaced across the domain, and the loss metrics are computed only on the permeability field, for which we do not have any direct information. 

For the remaining three time-dependent PDEs, we select the ratio of observed data points to be 0.3\%, 1\%, and 3\%, with locations randomly sampled. The three loss metrics we use, selected from~\citep{takamoto2022pdebench}, are the root mean squared error (RMSE), the normalized root mean squared error (nRMSE), and the RMSE of the conserved quantity (cRMSE). These metrics are computed in the unknown regions of the fields. Due to the size of the dataset, we select 1000 samples for each problem for benchmarking. We abbreviate the 2D compressible Navier-Stokes equations as CFD in the following sections.

\begin{table}[!htb]
\centering
\caption{Results of 1000 unseen samples for different PDEs. The diffusion models have an ensemble size of 25 and solved for 20 steps with the predictor-corrector scheme.}
\label{tab:validation_results}
\resizebox{\textwidth}{!}{ 
\begin{tabular}{c|c|c|c|c|c|c}
\toprule
\textbf{PDEs} & \textbf{Obs\%} & \textbf{Metric} & \multicolumn{3}{c|}{\textbf{Diffusion Model}} & \textbf{VT-UNET} \\ 
\cline{4-6} 
& & & \makecell{\textbf{Guided} \\ \textbf{Sampling}} & \textbf{CFG} & \makecell{\textbf{Cross}\\\textbf{Attention}} &  \\ 
\midrule
\multirow{9}{*}{\makecell{Shallow \\ water}} 
& \multirow{3}{*}{0.3\%}  & RMSE & $8.01 \times 10^{-3}$ & $4.20 \times 10^{-3}$ & \(\bm{3.76 \times 10^{-3}}\) & $4.18 \times 10^{-3}$ \\ 
& & nRMSE & $1.36 \times 10^{0}$ & $8.52 \times 10^{-1}$ & \(\bm{7.88 \times 10^{-1}}\) & $8.26 \times 10^{-1}$ \\ 
& & cRMSE & $7.06 \times 10^{-4}$ & $3.98 \times 10^{-4}$ & \(\bm{3.75 \times 10^{-4}}\) & $4.19 \times 10^{-4}$ \\ 
\cline{2-7}
& \multirow{3}{*}{1\%} & RMSE & $7.89 \times 10^{-3}$ & $3.46 \times 10^{-3}$ & $3.01 \times 10^{-3}$ & \(\bm{2.60 \times 10^{-3}}\) \\ 
& & nRMSE & $1.32 \times 10^{0}$ & $4.15 \times 10^{-1}$ & $\bm{3.47 \times 10^{-1}}$ & \(3.49 \times 10^{-1}\) \\ 
& & cRMSE & $6.96 \times 10^{-4}$ & $2.39 \times 10^{-4}$ & $2.30 \times 10^{-4}$ & \(\bm{2.08 \times 10^{-4}}\) \\ 
\cline{2-7}
& \multirow{3}{*}{3\%} & RMSE & $7.56 \times 10^{-3}$ & $3.21 \times 10^{-3}$ & $2.66 \times 10^{-3}$ & \(\bm{1.55 \times 10^{-3}}\) \\ 
& & nRMSE & $1.22 \times 10^{0}$ & $3.11 \times 10^{-1}$ & $2.56 \times 10^{-1}$ & \(\bm{1.81 \times 10^{-1}}\) \\ 
& & cRMSE & $6.46 \times 10^{-4}$ & $1.82 \times 10^{-4}$ & $1.70 \times 10^{-4}$ & \(\bm{1.16 \times 10^{-4}}\) \\ 
\midrule
\multirow{9}{*}{\makecell{Diffusion \\ reaction}} 
& \multirow{3}{*}{0.3\%} & RMSE & $7.94 \times 10^{-2}$ & $7.10 \times 10^{-2}$ & $6.19 \times 10^{-2}$ & \(\bm{6.09 \times 10^{-2}}\) \\ 
& & nRMSE & $9.94 \times 10^{-1}$ & $8.68 \times 10^{-1}$ & $7.53 \times 10^{-1}$ & \(\bm{7.41 \times 10^{-1}}\) \\ 
& & cRMSE & $1.80 \times 10^{-2}$ & $3.49 \times 10^{-3}$ & $\bm{3.07 \times 10^{-3}}$ & \(3.09 \times 10^{-3}\) \\ 
\cline{2-7}
& \multirow{3}{*}{1\%} & RMSE & $7.81 \times 10^{-2}$ & $5.93 \times 10^{-2}$ & $3.47 \times 10^{-2}$ & \(\bm{3.46 \times 10^{-2}}\) \\ 
& & nRMSE & $9.73 \times 10^{-1}$ & $7.09 \times 10^{-1}$ & $4.06 \times 10^{-1}$ & \(\bm{4.04 \times 10^{-1}}\) \\ 
& & cRMSE & $1.75 \times 10^{-2}$ & $2.03 \times 10^{-3}$ & $1.43 \times 10^{-3}$ & \(\bm{1.39 \times 10^{-3}}\) \\ 
\cline{2-7}
& \multirow{3}{*}{3\%} & RMSE & $7.50 \times 10^{-2}$ & $4.47 \times 10^{-2}$ & \(\bm{1.83 \times 10^{-2}}\) & $1.85 \times 10^{-2}$ \\ 
& & nRMSE & $9.19 \times 10^{-1}$ & $5.01 \times 10^{-1}$ & \(\bm{1.59 \times 10^{-1}}\) & $1.62 \times 10^{-1}$ \\ 
& & cRMSE & $1.65 \times 10^{-2}$ & $1.24 \times 10^{-3}$ & $7.95 \times 10^{-4}$ & \(\bm{7.39 \times 10^{-4}}\) \\ 
\midrule
\multirow{9}{*}{\makecell{CFD}} 
& \multirow{3}{*}{0.3\%} & RMSE & $5.53 \times 10^{0}$ & $2.48 \times 10^{-1}$ & \(2.01 \times 10^{-1}\) & $\bm{1.70 \times 10^{0}}$ \\ 
& & nRMSE & $2.46 \times 10^{0}$ & $2.23 \times 10^{-1}$ & \(\bm{1.28 \times 10^{-1}}\) & $1.42 \times 10^{-1}$ \\ 
& & cRMSE & $6.07 \times 10^{0}$ & $1.28 \times 10^{-1}$ & \(8.31 \times 10^{-2}\) & $\bm{4.92 \times 10^{-2}}$ \\ 
\cline{2-7}
& \multirow{3}{*}{1\%} & RMSE & $5.27 \times 10^{0}$ & $1.79 \times 10^{-1}$ & $1.24 \times 10^{-1}$ & \(\bm{8.38 \times 10^{-2}}\) \\ 
& & nRMSE & $2.36 \times 10^{0}$ & $1.47 \times 10^{-1}$ & $7.89 \times 10^{-2}$ & \(\bm{6.89 \times 10^{-2}}\) \\ 
& & cRMSE & $5.78 \times 10^{0}$ & $9.35 \times 10^{-2}$ & $6.78 \times 10^{-2}$ & \(\bm{1.86 \times 10^{-2}}\) \\ 
\cline{2-7}
& \multirow{3}{*}{3\%} & RMSE & $4.57 \times 10^{0}$ & $1.33 \times 10^{-1}$ & $7.66 \times 10^{-2}$ & \(\bm{4.27 \times 10^{-2}}\) \\ 
& & nRMSE & $2.08 \times 10^{0}$ & $1.03 \times 10^{-1}$ & $5.77 \times 10^{-2}$ & \(\bm{3.71 \times 10^{-1}}\) \\ 
& & cRMSE & $5.01 \times 10^{0}$ & $7.11 \times 10^{-2}$ & $5.26 \times 10^{-2}$ & \(\bm{1.02 \times 10^{-2}}\) \\ 
\midrule
\multirow{6}{*}{Darcy} 
& \multirow{3}{*}{0.3\%} & RMSE & $6.43 \times 10^{-1}$ & $3.91 \times 10^{-1}$ & $2.36 \times 10^{-1}$ & \(\bm{2.34 \times 10^{-1}}\) \\ 
& & nRMSE & $4.76 \times 10^{-1}$ & $2.91 \times 10^{-1}$ & $1.78 \times 10^{-1}$ & \(\bm{1.76 \times 10^{-1}}\) \\ 
& & cRMSE & $7.80 \times 10^{-2}$ & $3.54 \times 10^{-2}$ & $1.38 \times 10^{-2}$ & \(\bm{9.91 \times 10^{-3}}\) \\ 
\cline{2-7}
& \multirow{3}{*}{1.37\%} & RMSE & $6.40 \times 10^{-1}$ & $3.48 \times 10^{-1}$ & $1.74 \times 10^{-1}$ & \(\bm{1.25 \times 10^{-1}}\) \\ 
& & nRMSE & $4.74 \times 10^{-1}$ & $2.61 \times 10^{-1}$ & $1.29 \times 10^{-1}$ & \(\bm{9.18 \times 10^{-2}}\) \\ 
& & cRMSE & $7.98 \times 10^{-2}$ & $2.91 \times 10^{-2}$ & $1.92 \times 10^{-2}$ & \(\bm{7.31 \times 10^{-3}}\) \\ 
\bottomrule
\end{tabular}
}
\end{table}

\begin{figure}[!htb]
    \centering
    \includegraphics[width=0.9\textwidth]{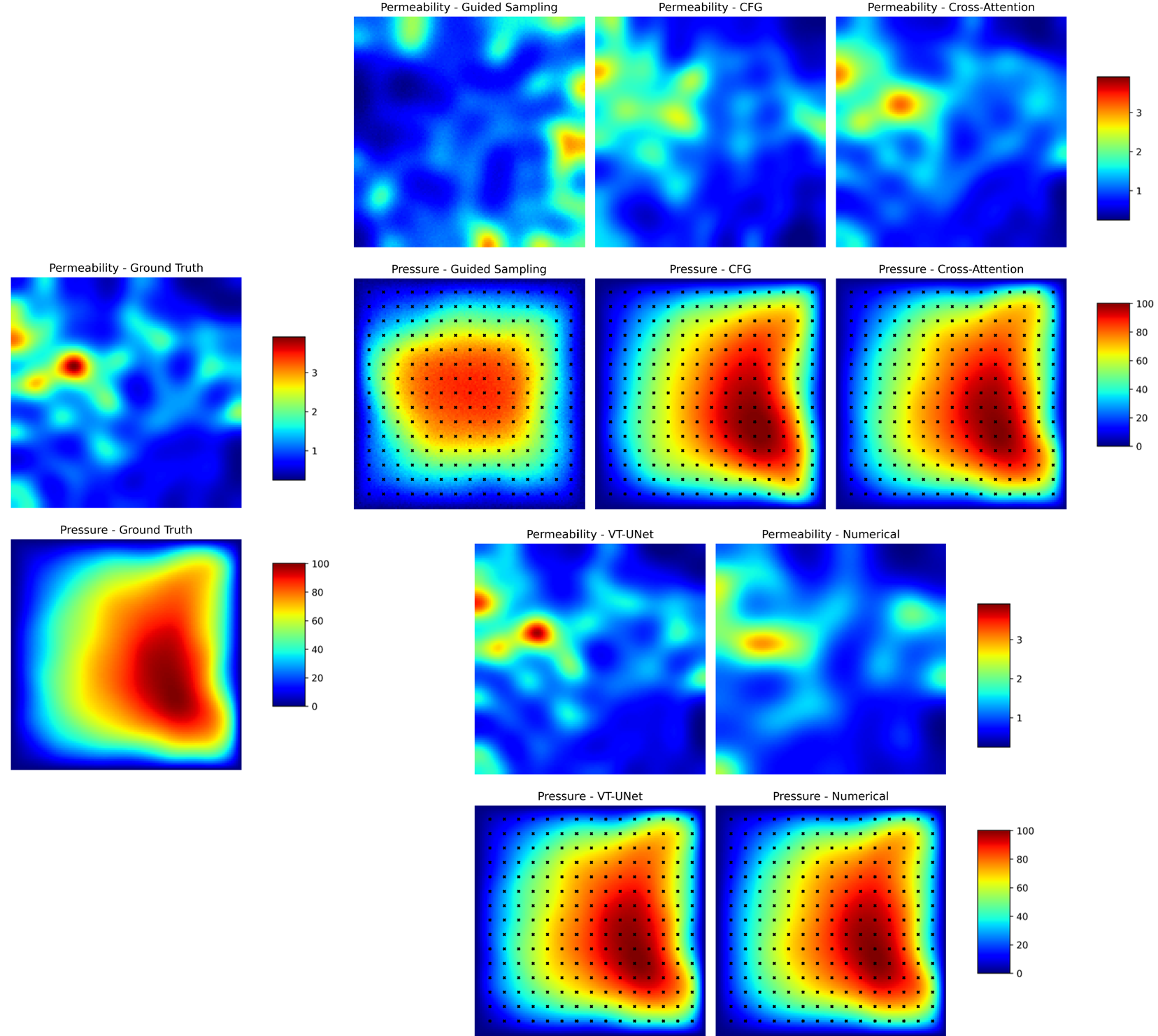}
    \caption{Comparison of the generated permeability fields for the Darcy flow problem with 1.37\% observed data points. Reverse sampling process of the diffusion models is configured with 20 steps, using a predictor-corrector scheme and a single trajectory. The black crosses denote the observed data points.}
    \label{fig:darcy_compare}
\end{figure}

\begin{table}[!htb]
\centering
\caption{Comparison of nRMSE and Computation Cost per sample for the Darcy flow, the computation cost of diffusion models are computed from an ensemble of 25 trajectories with predictor-corrector and 20 steps.}
\label{tab:darcy_comparison_results}
\begin{tabular}{c c c c c c}
\toprule
 & \makecell{\textbf{Guided} \\ \textbf{sampling}} & \textbf{CFG} & \makecell{\textbf{Cross-}\\ \textbf{Attention}} & \textbf{VT-UNet} & \textbf{Numerical} \\ 
\midrule
nRMSE (0.3\%) & 0.476 & 0.291 & 0.178 & 0.176 & 0.202 \\ 
nRMSE (1.37\%) & 0.474 & 0.261 & 0.129 & 0.092 & 0.180 \\ 
\midrule
Computation cost (s) & 0.944 & 0.931 & 1.769 & 0.00206 & 62 \\ 
\bottomrule
\end{tabular}
\end{table}

The reconstructed fields from noiseless observations using different methods are compared in Table~\ref{tab:validation_results}. The results for the diffusion models are generated from an ensemble of 25 trajectories using the predictor-corrector scheme with 20 steps. We found that the predictor-corrector scheme generally provides more robust reconstructions than the multistep solver~\cite{lu2022dpm}; reconstructions from the multistep solver are included in ~\ref{app:multistep}. Among the different conditioning methods, cross-attention consistently shows the best performance in terms of RMSE, nRMSE, and cRMSE across all PDEs.

With the same number of training steps, VT-UNet with noiseless observations achieves the best performance in nRMSE for 39 out of 43 cases. This could be caused by the diffusion model has an additional implicit dimension to learn, specifically, the noise level, which makes the optimization problem more challenging. Additionally, diffusion models tend to have higher cRMSE values than VT-UNet, even when the computed nRMSE is lower. This may be attributed to the architecture of the EDM formulation, where the model's output is suppressed in the low-noise region (Equation ~\eqref{eq:edm_loss}), and the log-normal noise schedule primarily focuses on the medium-noise region, leaving the model less capable of correcting fine details in the low-noise region.

\begin{figure}[!htb]
    \centering
    \includegraphics[width=\textwidth]{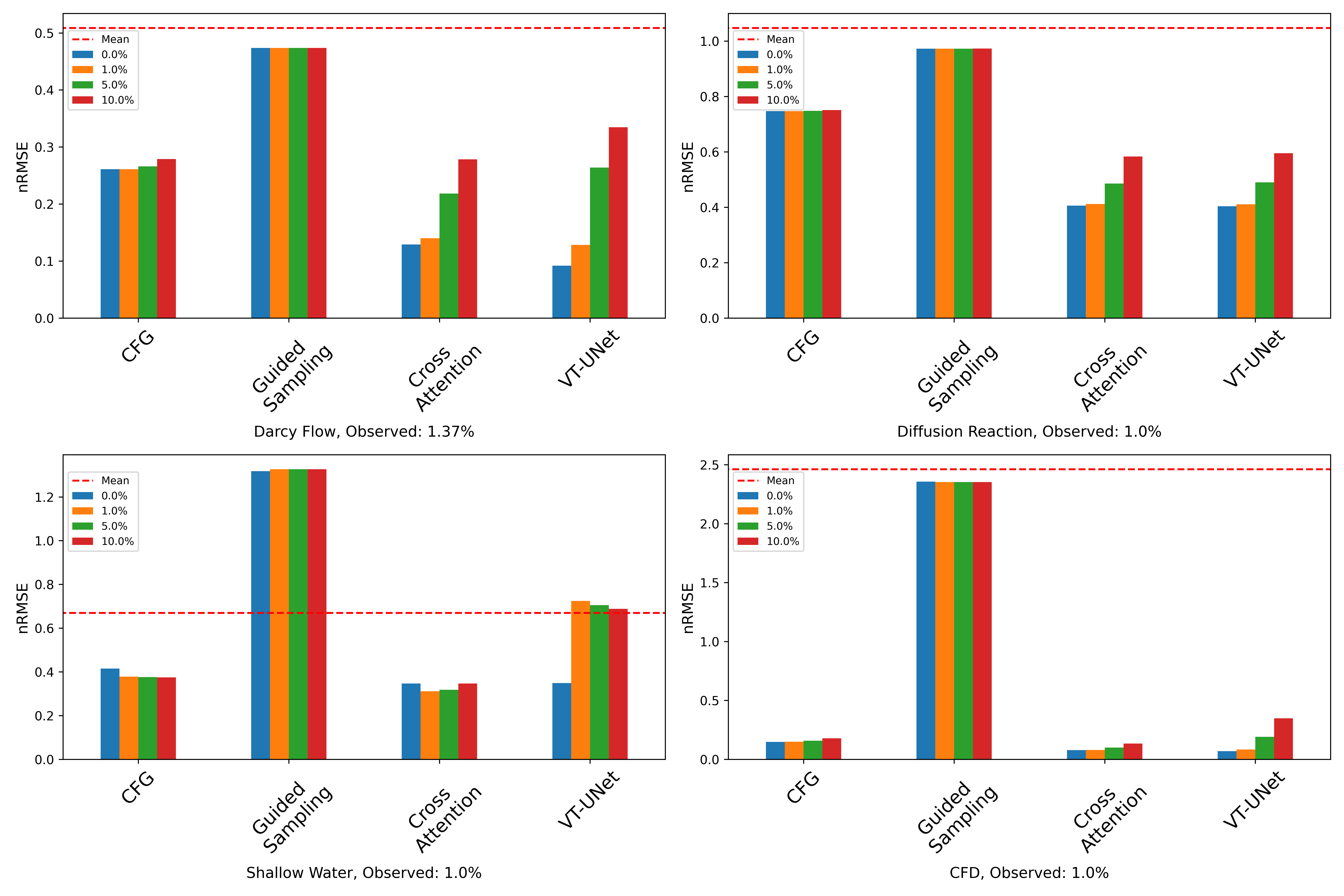}
    \caption{Bar plot of nRMSE for the PDEs with 1\% observed data points (1.37\% for the Darcy flow) and various observation noise levels. The red dashed line denotes the error of reconstructing the field using the mean of the training data. The diffusion models are configured with 20 steps, with a predictor-corrector scheme and an ensemble of 25 trajectories.}
    \label{fig:bar_chart_0.01}
\end{figure}

However, VT-UNet is more sensitive to observation noise levels, as shown in Figure ~\ref{fig:bar_chart_0.01}. The diffusion models demonstrate more stable performance across different noise levels. When noise levels increase to $5\%$, the cross-attention method outperforms VT-UNet in all cases except for the CFD problem. The cross-attention method generally outperforms other conditioning methods across all observation noise levels. In contrast, the guided sampling method shows the worst performance for all PDEs and observation noise levels, indicating that for complex physical systems with sparse observations, guided sampling is insufficient to steer the sampling trajectory toward the correct solution. A comparison of the diffusion-reaction problem under colored noise conditions is provided in~\ref{app:color_noise}. If unconditional generation capability is also desired, one should consider using the ControlNet approach~\cite{jacobsen2023cocogen} or setting the guidance scale in Equation~\eqref{eq:cfg} to be less than one.

\begin{figure}[!htb]
    \centering
    \includegraphics[width=\textwidth]{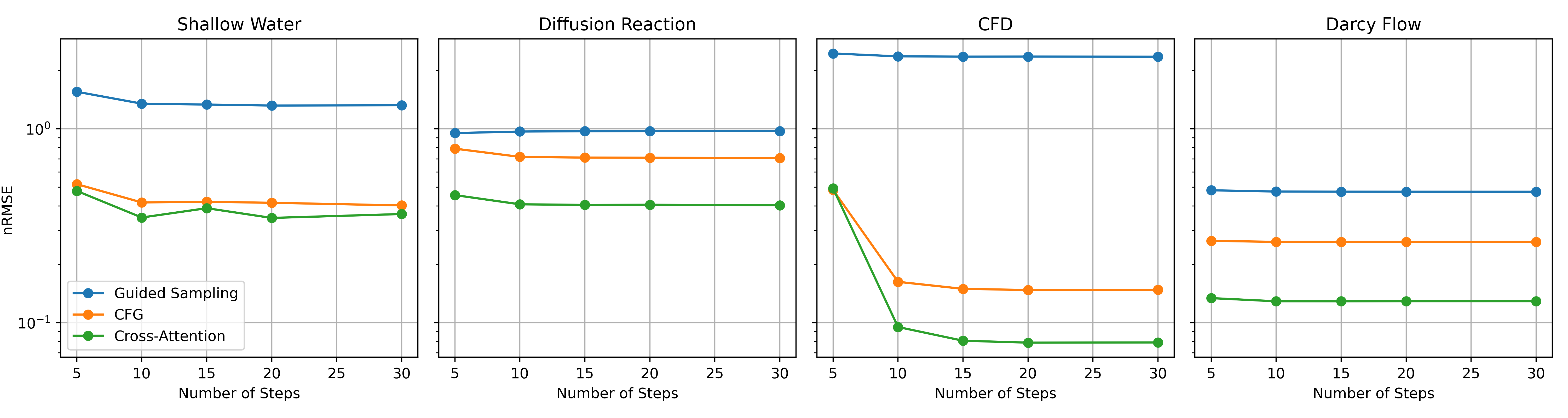}
    \caption{nRMSE of the PDEs with 1\% observed data points (1.37\% for the Darcy flow) for different numbers of reverse steps. The diffusion models are configured with a predictor-corrector scheme and an ensemble of 25 trajectories.}
    \label{fig:step_sweep}
\end{figure}

We also investigate the effect of the number of reverse steps on the performance of the diffusion models. The results are shown in Figure ~\ref{fig:step_sweep}. The performance of the diffusion models generally improves as the number of reverse steps increases, and we do not observe a turning point where the performance starts to degrade. However, this improvement is only significant in the CFD problem, suggesting that the learned mapping trajectories between the data distribution and the Gaussian prior for the other PDEs are less complex. This is because the reverse path solved by the PF ODE is only an approximation of the continuous reverse path. If comparable performance can be achieved with fewer reverse steps, it indicates that the flow path has a high degree of 'straightness'~\citep{liu2022flow}. Additionally, the improvement diminishes after a certain number of steps, with 20 steps providing a good trade-off between performance and computational cost.

We compare the nRMSE of the deep learning methods to that of the numerical iterative Kalman filtering approach on the Darcy Flow problem, where sparse measurements of the pressure field are provided to reconstruct two fields. The evaluations on the reconstructed permeability fields are shown in Table~\ref{tab:darcy_comparison_results}. Computation time is calculated as the average time required to infer a batch of reconstructed fields. For the deep learning models, the time is measured on a Nvidia H100 GPU, while for the numerical approach, the time is measured on an Intel 13700K CPU. We set the number of KLE modes for the numerical method to 64, using the recommended regularization hyperparameter of 0.5~\cite{huang2022iterated}. The diffusion models with cross-attention and the VT-UNet both outperform the numerical approach in terms of nRMSE and computational cost. For the other time-dependent PDEs, the numerical approach is not applicable due to its high computational cost.

\begin{figure}[!htb]
    \centering
    \subfigure[Reconstructed fields that close to the start of the simulation (step: 11/101).
    ]{\includegraphics[width=0.9\textwidth]{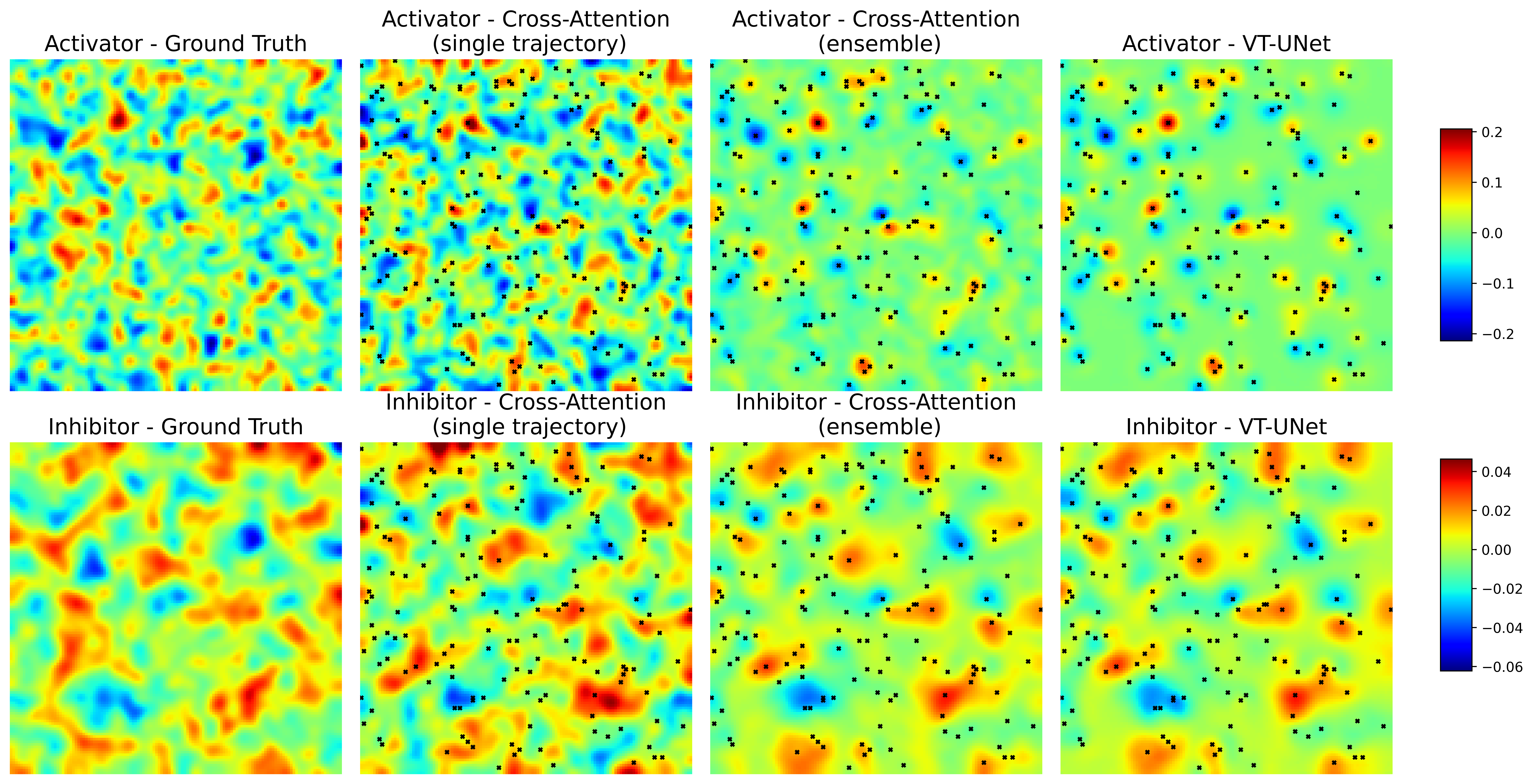}}
    \vspace{1em} %
    \subfigure[Reconstructed fields that close to the end of the simulation (step: 91/101).
    ]{\includegraphics[width=0.9\textwidth]{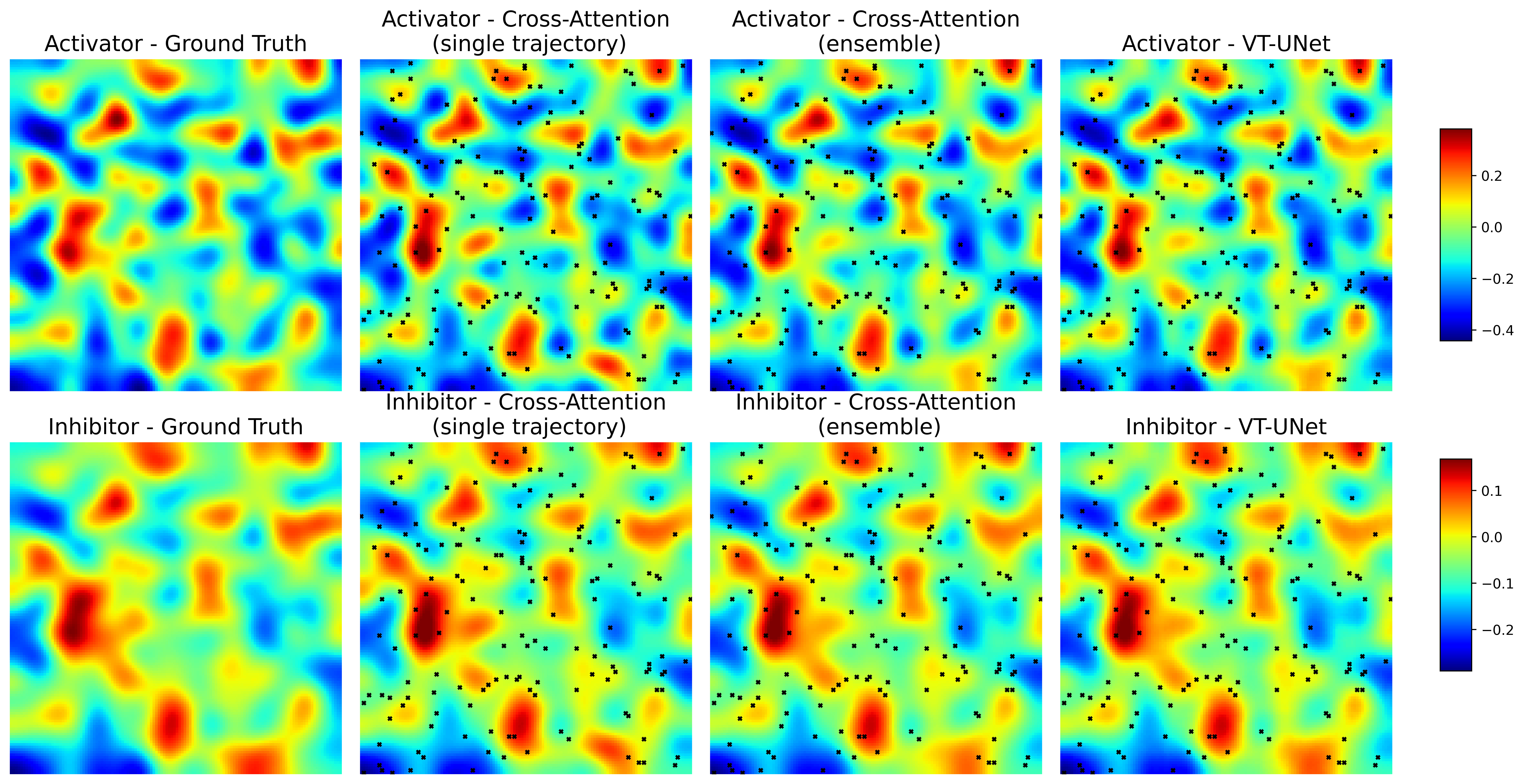}}
    \caption{Comparison of the generated fields by VT-UNet, single trajectory and ensemble mean of cross-attention diffusion model for the Diffusion Reaction equations with 1\% observed data points. The diffusion models are configured with 20 steps, with a predictor-corrector scheme and an ensemble of 25 trajectories. The black crosses denote the observed data points.}
    \label{fig:dr_compare}
\end{figure}

Field reconstruction tasks from sparse observations are underdetermined problems, meaning that multiple solutions can exist for the same set of measurements. This is best illustrated in the Diffusion-Reaction equations, where the concentration profiles of the activator and inhibitor fields evolve from high-frequency noise to smooth patterns, as shown in Figure ~\ref{fig:dr_compare}.

At the high-frequency stage, the VT-UNet fails to capture the possible details in the fields, although the mean representation has a lower MSE error. This outcome is expected since the training objective is formulated as an MSE loss. The diffusion models, on the other hand, are able to capture the high-frequency patterns in the fields and provide a possible realization of the observations. This approach offers a new perspective on understanding the possible underlying structures of the fields, though it typically results in a larger error compared to the deterministic mean representation. The capability of generating different outcomes was also reported in~\cite{haitsiukevich2024diffusion}. However, the mean derived from an ensemble of reverse-sampled trajectories with the same observation points also converges to a similar mean representation. This indicates that the fields reconstructed by the diffusion models are consistent with the results obtained from the deterministic method.

\begin{figure}[!htb]
    \centering
    \includegraphics[width=\textwidth]{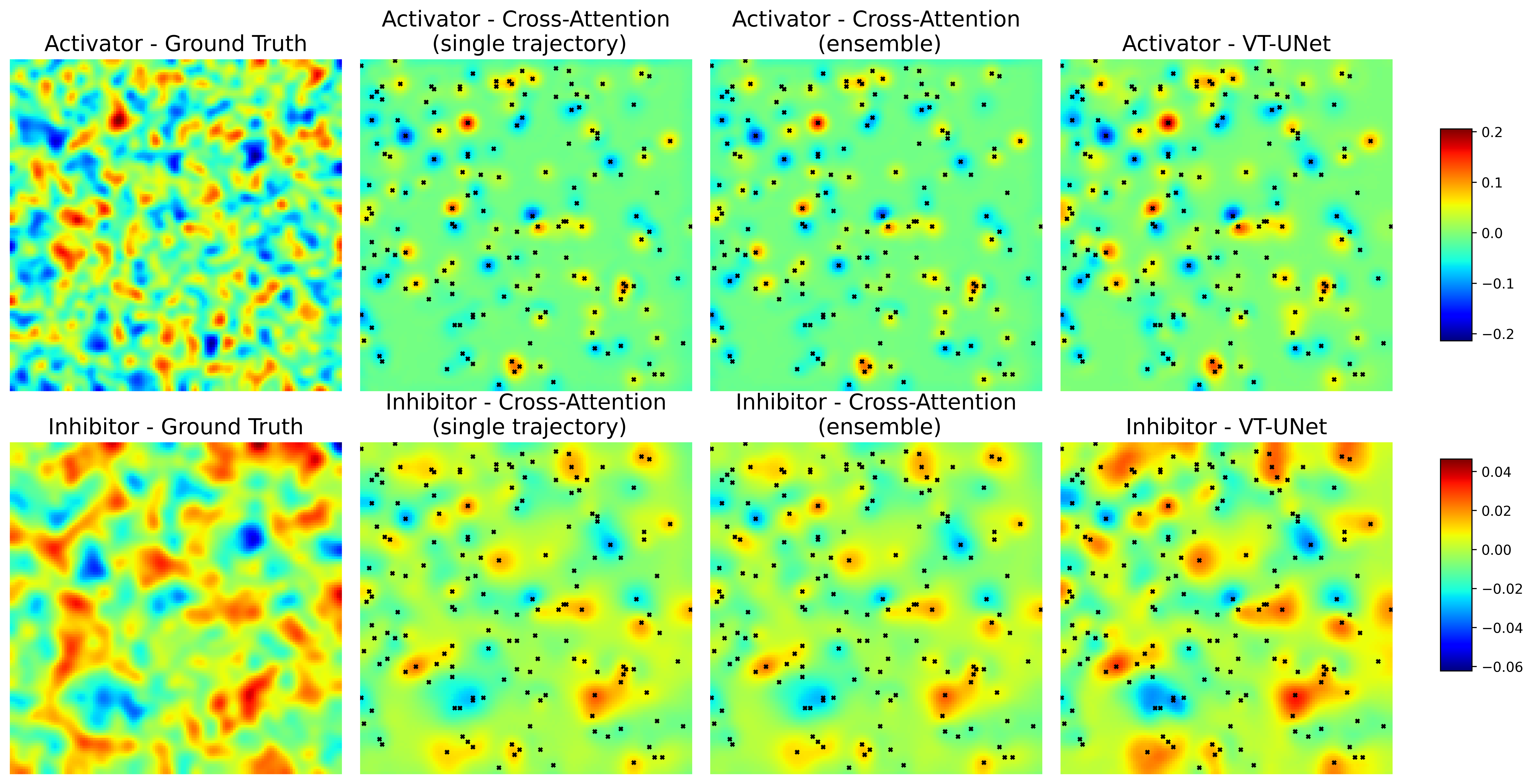}
    \caption{Comparison of the generated fields by VT-UNet, single trajectory and ensemble mean of cross-attention diffusion model for the Diffusion Reaction equations with 1\% observed data points. The diffusion models are configured with 20 steps, with a multistep scheme and an ensemble of 25 trajectories.}
    \label{fig:dr_compare_multistep}
\end{figure}

In the low-frequency stage, both the VT-UNet and the diffusion models effectively capture the underlying structure of the fields, with the difference between a single trajectory and the ensemble mean of the reconstructed fields being less significant. It is also noted that when the fields are sampled using the multistep solver, the diffusion models lose the ability to capture possible realizations, as shown in Figure~\ref{fig:dr_compare_multistep}. A comparison between different conditioning methods using multistep sampling is provided in~\ref{app:multistep}. The errors associated with multistep~\cite{lu2022dpm} sampling are generally higher than those of the predictor-corrector method, although multistep sampling achieves better nRMSE for the fine-tuned model (see~\ref{app:res}). Therefore, if a mean representation is desired, using the multistep solver can reduce computational cost with only a slight increase in error.

\begin{figure}[!htb]
\centering
\includegraphics[width=\textwidth]{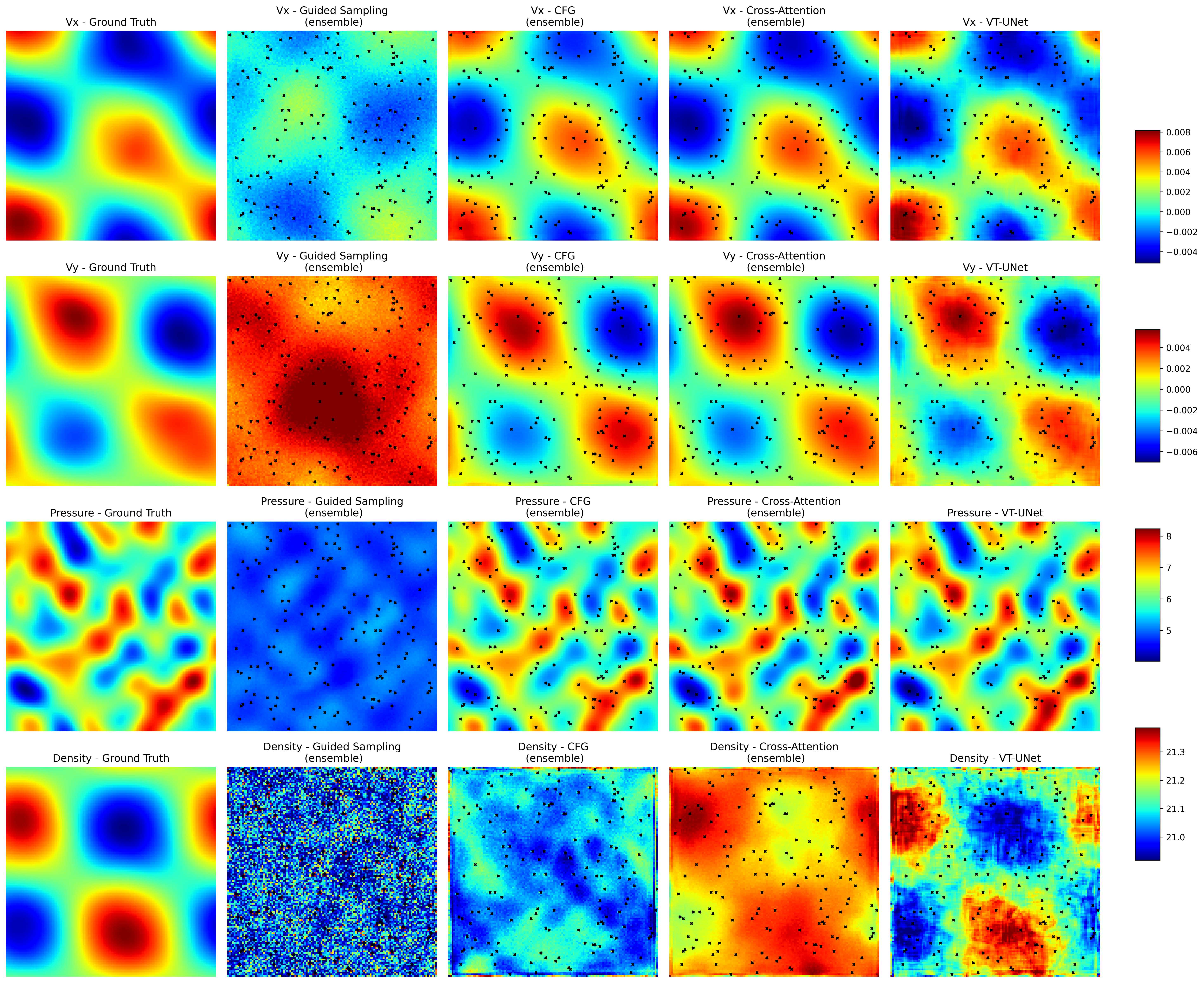}
\caption{Comparison of the generated fields by VT-UNet and the ensemble means of diffusion models for the compressible Navier-Stokes equations with 1\% observed data points. The diffusion models are configured with 20 steps, using a predictor-corrector scheme and an ensemble of 25 trajectories.}
\label{fig:ns_compare}
\end{figure}

The results for the compressible Navier-Stokes equations are shown in Figure ~\ref{fig:ns_compare}. The diffusion models with CFG and cross-attention provide better reconstructions of the velocity fields compared to VT-UNet. However, for the density field, all methods fail to accurately capture the interface, despite the small relative error. This may be attributed to the heavy-tailed distribution of the density field, as shown in Figure ~\ref{fig:ns_hist}, which is not effectively handled by the normalization during data preprocessing.

\begin{figure}[!htb]
\centering
\includegraphics[width=0.88\textwidth]{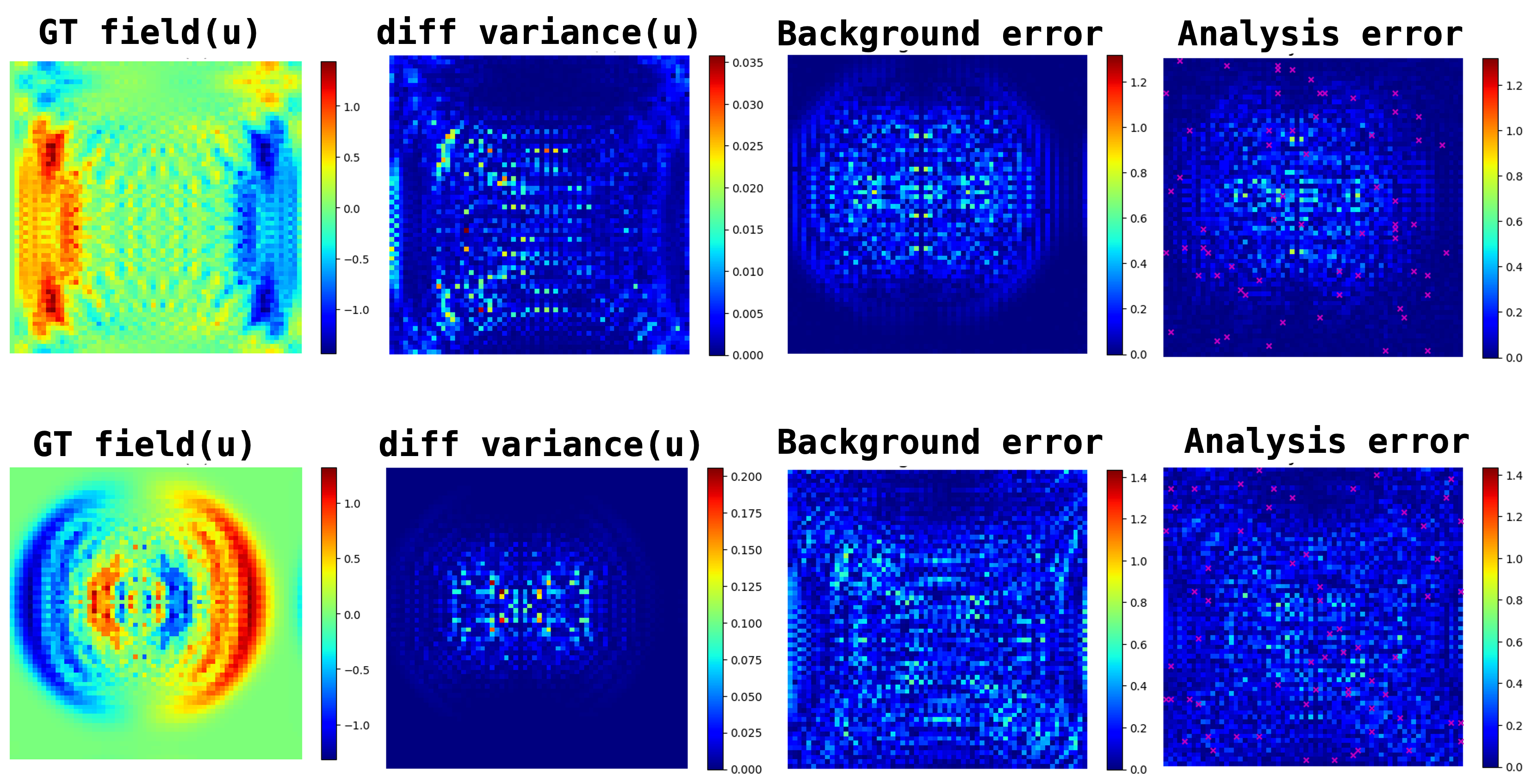}
\caption{Samples of prior (background) and posterior (analysis) error of data assimilation applied on the 2D shallow water test case at time steps 20 (first row) and 40 (second row). Red dots illustrate the observation positions.} 
\label{fig:DA_compare}
\end{figure}

Figure~\ref{fig:DA_compare} displays the assimilated velocity field in the shallow water application following the diffusion model reconstruction. As mentioned in Section~\ref{sec:DA}, the background error covariance in this case is empirically estimated from the ensemble generated using the diffusion model introduced in this paper. The ensemble size is fixed at 10 for all DA experiments.
As can be clearly observed in Figure~\ref{fig:DA_compare}, the field reconstruction error is significantly reduced posterior to the DA process, particularly around the observable points. The estimated variance (i.e., the diagonal of the covariance matrix estimated using the 10 realizations of the diffusion model) is also shown in Figure~\ref{fig:DA_compare}. An energy plot of the shallow water equation is provided in~\ref{app:energy}, demonstrating that the models maintain consistent performance over time, even when each time step is treated as a separate instance.

\begin{figure}[!htb]
\centering
\includegraphics[width=0.65\textwidth]{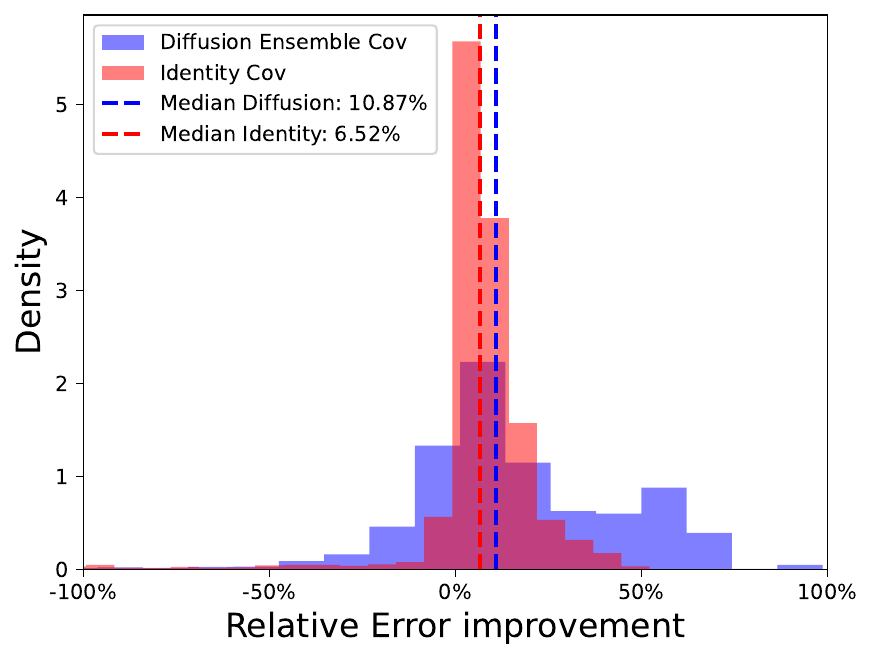}
\caption{Histogram of relative error improvement $Im_{t}$ distribution with different DA error covariances. }
\label{fig:ns_compare_curve}
\end{figure}

As a benchmark, we also conducted DA using an identity background error covariance, which is a common choice in practical DA when $\textbf{B}_{\Tilde{t}}$ cannot be explicitly specified. Numerical experiments are repeated for all 25 simulations in the test datasets of the shallow water simulations. We calculate the relative error improvement $Im_{t}$, defined as:

\begin{align}
     Im_{t} = \frac{||\bm{x}_{b,\Tilde{t}} - \bm{x}_{\textrm{true},\Tilde{t}}||_2 - ||\bm{x}_{a,\Tilde{t}} - \bm{x}_{\textrm{true},\Tilde{t}}||_2}{||\bm{x}_{b,\Tilde{t}} - \bm{x}_{\textrm{true},\Tilde{t}}||_2},
 \end{align}

which represents the improvement in field reconstruction due to the DA process.The distribution of $Im_{t}$ in the test dataset, consisting of 50 simulations (each evaluated on 10 samples), is presented in Figure~\ref{fig:ns_compare_curve}. Overall, both DA methods using either the diffusion ensemble covariance matrix or the identity covariance matrix, improve the average field reconstruction accuracy. However, the diffusion ensemble covariance matrix demonstrates superior performance in most of the corrections applied to the diffusion model output.
It is important to note that the placement of observable points varies over time at different time steps. In some cases, DA may result in negative improvement due to the sparsity of observations and potential overfitting by the PCA algorithm.
Overall, these results demonstrate that the proposed diffusion model can seamlessly incorporate a Bayesian fine-tuning method such as DA, to further enhance the accuracy of field reconstruction.

\section{Conclusion} \label{sec:conclusion}

We enhance and evaluate diffusion models for field reconstruction tasks, with the goal of  estimating complete spatio-temporal fields from sparse observations. By introducing a novel condition encoding block that integrates Voronoi-tessellated fields and sensing positions as an inductive bias, we constructed a tractable mapping between observed and unobserved regions. This approach leverages Feature-wise Linear Modulation (FiLM) and self-attention mechanisms to effectively capture the conditioning representation and support probabilistic reconstruction. We benchmark the effectiveness of conditioning using two commonly employed methods: hidden state augmentation, which we refer to as classifier-guidance free (CFG), and the cross-attention mechanism, against the adapted deterministic method, VT-UNet, with the same number of training steps. In addition, we include guided sampling in our comparison, a commonly used method that operates in the reverse sampling process without requiring explicit conditioning.


The proposed conditional encoding is shown to enable the diffusion model to generate high-quality fields from sparse observations. It offers a flexible approach to handle time-dependent PDEs without the need for explicit physical time conditioning, making it particularly effective in scenarios involving moving sensors. Our benchmarks for model evaluations includes Darcy flow, shallow water equations, diffusion-reaction equations, and compressible Navier-Stokes equations.

Our numerical experiments  show that in the steady state Darcy flow problem, the diffusion model outperforms traditional numerical iterative method in terms of accuracy and computational efficiency. Although the diffusion model does not surpass the interpolation-based deterministic model in noiseless settings with the same training effort due to the added complexity of learning across various noise levels, it proves to be more robust under noisy observations, which is critical for real-world applications. As the number of variables and the resolution of the domain increase, the difficulty of training the full-field diffusion model is expected to rise significantly, emphasizing the need for implementing latent diffusion models for high-dimensional problems.

Among the tested conditioning methods, the cross-attention mechanism within the condition encoding block generally provides the best performance. Conversely, the guided sampling method fails to reconstruct the correct fields for all PDEs. Regarding the different PF ODE solvers for the reverse sampling process, we found that the predictor-corrector scheme is more robust than the multistep scheme on the EDM framework, as it able to capture possible realizations of the underdetermined reconstruction with sparse observations. Furthermore, we demonstrate that the mean of these realizations converges to the output obtained by the deterministic model, suggesting that the encoding block effectively extracts information from the inductive bias and sensing positions. While our tests focus on non-periodic dynamics, we expect the diffusion model to also perform well on periodic problems, similar to findings from previous work on VCNN~\cite{fukami2021global}.

Additionally, our experiments indicate that  data assimilation methods can be integrated with the proposed diffusion model to further improve accuracy. The stochastic nature of the diffusion model can also aid in uncertainty quantification in inverse modeling through an ensemble approach, as demonstrated in this study. In future work, we plan to further explore the integration of diffusion models within the ensemble data assimilation framework for high-dimensional dynamical systems.



\section*{Data availability}

All the data used are publicly available or can be generated from publicly available code. The source code for the experiments is available on GitHub:
\newline
\href{https://github.com/tonyzyl/DiffusionReconstruct}{https://github.com/tonyzyl/DiffusionReconstruct}.

\section*{Acknowledgment}
This work was supported by Los Alamos National Laboratory under the project “Algorithm/Software/Hardware Co-design for High Energy Density applications” at the University of Michigan. Sibo Cheng acknowledges the support of the French Agence Nationale de la Recherche (ANR) under reference ANR-22-CPJ2-0143-01.

\bibliographystyle{elsarticle-num}
\bibliography{references_abbrev}  






\appendix
\section{Additional Information}

\subsection{Implementation and Training details \label{app:train}}

The hyperparameters of the EDM framework are designed to handle normally distributed data with a standard deviation of 0.5. Therefore, we scale our data accordingly to achieve a similar distribution based on the mean and standard deviation of the training data.

The U-Net architecture is utilized for the denoiser function, $D_\theta$, following the same design as in~\citep{karras2022elucidating}. The implementation is based on the PyTorch 2.3.1 and diffusers 0.29.2 libraries. The network consists of $N_{block}$ down-sampling and up-sampling blocks, where $N_{block}$ is the number of blocks required to achieve a bottleneck size of $16 \times 16$, where $N_{block}$ is the number of blocks required to achieve a bottleneck size of $16 \times 16$, the last block in the down-sampling part and the first block in the up-sampling part do not have down-sampling and up-sampling layers, respectively. Specifically, $N_{block}$ equals 3 for the shallow water equations and 4 for the other problems. The first down-sampling block has 128 channels, while the remaining blocks have 256 channels. The up-sampling blocks are symmetric to the down-sampling blocks. The network uses a 3x3 convolutional layer with a stride of 2 for down-sampling and a nearest interpolation followed by a 3x3 convolutional layer for up-sampling. FiLM~\cite{perez2017film} is applied for both the noise level embedding and the classifier-free guidance (CFG).

Zero padding in CNNs can leak positional information~\cite{islam2020much}, and typically, special treatment is required for processing different types of periodic conditions~\cite{morimoto2021convolutional}. In this work, we do not implement these techniques; instead, we rely on skip connections and self-attention mechanisms to capture global information. However, we expect that periodic padding could improve the generalizability of the models and could be explored in future work.

The network is trained using the AdamW optimizer with a learning rate of $10^{-4}$ and a weight decay of $0.01$ for 100,000 steps, with a batch size of 128 samples per step, on 8 Nvidia H100 GPUs. The loss is calculated using the MSE between the noiseless fields and the denoised fields, with weighting as proposed in~\citep{karras2022elucidating}. The model weights are updated using an EMA with a decay rate of 0.999, $\text{inv\_gamma}=1.0$, and $\text{power}=0.75$. The training data are selected from the first 80\% segment of the dataset, and we do not save the best weights based on validation.

Based on Equation~\eqref{eq:score_partial_obs}, the observed region is randomly sampled from the field solution with a ratio drawn from a $\mathcal{U}(0, 0.1)$ distribution, and the observation is merged with random noise in the unobserved regions according to the noise schedule. For the time-dependent PDEs, snapshots from each simulation are unraveled during training, and physical time is not provided as conditioning information.

\subsection{Noise Schedule \label{app:noise}}

Various noise schedulers exist for diffusion models~\cite{esser2024scaling}, with the log-normal noise scheduler first proposed in the EDM framework~\cite{karras2022elucidating} using parameters $P_{\text{mean}} = -1.2$ and $P_{\text{std}} = 1.2$. It has been shown that the log-normal noise schedule needs to be tuned for optimal performance~\cite{karras2024analyzing}. In our experiments, we found that the noise schedule used during training (Equation~\eqref{eq:edm_traj}) should provide sufficient coverage of high noise levels to account for the large variability in physical fields. This ensures that the model encounters enough examples where the added noise is large enough to cause the field values to approximate a Gaussian distribution. Without this coverage, the diffusion model may struggle to generate correct fields from the initial Gaussian prior, even if it can effectively denoise the fields when the noise level is low. In our testing, we found that setting $P_{\text{mean}} = 1.2$ and $P_{\text{std}} = 1.7$ is robust for our tasks.

\subsection{Reduced order data assimilation \label{sec:app_DA}}
Conducting DA in the complete physical space can be both computationally intensive and time-consuming due to the high dimensionality of the state space. Additionally, when the state and observation spaces overlap (e.g., both are sampled from the velocity field), performing DA in the full physical space without carefully tuning the error covariance matrices may result in only point-to-point correlations.
In this section, we describe how the proposed method can be combined with a ROM using PCA to enhance efficiency further.

Given a set of $n_\textrm{state}$ state snapshots obtained from one or more simulations or predictions, these snapshots are organized into a matrix $\mathcal{X} \in \mathbb{R}^{(N_d \times N_d) \times n_\textrm{state}}$. In this matrix, each column corresponds to a flattened state at a specific time step, expressed as:
\begin{align}
\mathcal{X} = \big[ \bx_0 \big| \bx_1 \big| \dots \big| \bx_{n_\textrm{state}-1} \big].
\end{align}

The empirical covariance matrix $\bC_{\mathcal{X}}$ associated with $\mathcal{X}$ can be computed and expressed as:

\begin{align}
    \bC_{\mathcal{X}} = \frac{1}{n_\textrm{state}-1} \mathcal{X} \mathcal{X}^T = {\bL}_{\mathcal{X}} {\bD}_{\mathcal{X}} {{\bL}_{\mathcal{X}}}^T \label{eq:C}
\end{align}

Here, the columns of ${\bL}_{\mathcal{X}}$ represent the principal components of $\mathcal{X}$, and ${\bD}_{\mathcal{X}}$ is a diagonal matrix containing the corresponding eigenvalues ${ \lambda_{\mathcal{X},i}, i=0,\dots,n_\textrm{state}-1}$ arranged in descending order:
\begin{align}
  {\bD}_{\mathcal{X}} =
  \begin{bmatrix}
    \lambda_{\mathcal{X},0} & & \\
    & \ddots & \\
    & & \lambda_{\mathcal{X},n_\textrm{state}-1}
  \end{bmatrix}.
\end{align}

To reduce the dimensionality of the state variables to a space of dimension $q \hspace{2mm} (q\in \mathbb{N}^+ \hspace{2mm} \textrm{and} \hspace{2mm} q \leq n_\textrm{state})$, we derive a projection operator ${\bL}_{\mathcal{X},q}$ by selecting the first $q$ columns from ${\bL}_{\mathcal{X}}$. The matrix ${\bL}_{\mathcal{X}}$ can be obtained through Singular Value Decomposition (SVD), eliminating the need to estimate the full covariance matrix $\bC_{\mathcal{X}}$.

For a flattened state field $\bx_{\Tilde{t}}$, the reduced latent vector $\hat{\bx}_{\Tilde{t}}$ is calculated as:
\begin{align}
\hat{\bx}_{\Tilde{t}} = {{\bL}_{\mathcal{X},q}}^T \bx_{\Tilde{t}}, \label{eq: reconstruction}
\end{align}
which serves as an approximation to the complete vector $\bx_{\Tilde{t}}$.

This latent vector $\hat{\bx}_{\Tilde{t}}$ can then be expanded back to the full space vector $\bx_{\Tilde{t}}^r$ by:
\begin{align}
\bx_{\Tilde{t}}^r = {{\bL}_{\mathcal{X},q}} \hat{\bx}_{\Tilde{t}} = {{\bL}_{\mathcal{X},q}} ({{\bL}_{\mathcal{X},q}})^T \bx_{\Tilde{t}}.
\end{align}

 The assimilation process can be performed in the space of $\hat{\bx}_{\Tilde{t}}$ rather than in $\bx_{\Tilde{t}}$, resulting in a new state-observation operator $\hat{\mathcal{H}}$, which is defined as:
\begin{align}
\hat{\mathcal{H}} = \mathcal{H} \circ {{\bL}_{\mathcal{X},q}} \quad \textrm{with} \quad \by_{\Tilde{t}} = \mathcal{H}(\bx_{\Tilde{t}}) = \mathcal{H} \circ {{\bL}_{\mathcal{X},q}}(\hat{\bx}_{\Tilde{t}}) = \hat{\mathcal{H}}(\hat{\bx}_{\Tilde{t}}).
\end{align}

Thus the background error covariance matrix $\hat{\textbf{B}}_{\Tilde{t}}$ associated to $(\hat{\bm{x}},\hat{\mathcal{H}})$ can be obtained by
\begin{align}
   \hat{\textbf{B}}_{\Tilde{t}} =  {{\bL}_{\mathcal{X},q}}^T \textbf{B}_{\Tilde{t}} {\bL}_{\mathcal{X},q}
\end{align}.
As mentioned in section~\ref{sec:DA}, the $\textbf{B}_{\Tilde{t}}$ is estimated online using the generated diffusion ensemble.
Since the observation operator $\mathcal{H}$ and $\hat{\mathcal{H}}$ are linear, the DA here (Equation~\eqref{eq_3dvar}) is performed using the Best Linear Unbiased Estimator (BLUE) with the python-based ADAO~\cite{adaotrue} package.

\section{Datasets \label{app:data}}

\subsection{Darcy Flow}

\begin{figure}[!htb]
    \centering
    \includegraphics[width=\textwidth]{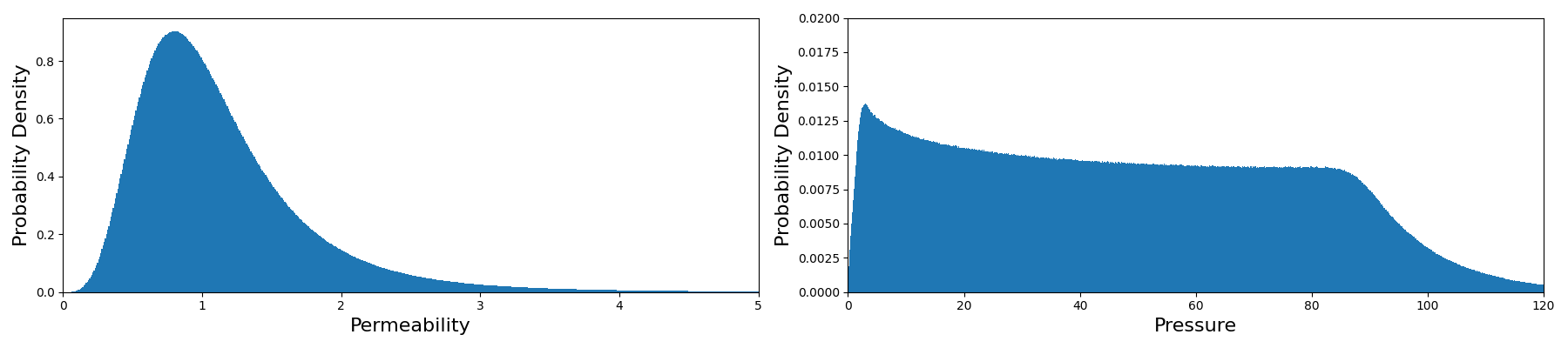}
    \caption{Histogram of the cell values of the permeability fields and the pressure fields of the Darcy flow data.}
    \label{fig:darcy_hist}
\end{figure}

We follow the formulations provided in~\cite{huang2022iterated}, the source function is defined as:
\begin{equation}
f(x_1, x_2) = \begin{cases} 
1000 & \text{if } 0 \leq x_2 \leq \frac{4}{6} \\
2000 & \text{if } \frac{4}{6} < x_2 \leq \frac{5}{6} \\
3000 & \text{if } \frac{5}{6} < x_2 \leq 1 
\end{cases}
\end{equation}
The generative parameter, $\theta$, is feed to the following KLE of the Gaussian field:
\begin{equation}
\log \alpha(\mathbf{x}, \theta) = \sum_{\mathbf{l}\in \mathbb{Z}^{0+} \times \mathbb{Z}^{0+}} \theta_{(\mathbf{l})} \sqrt{\lambda_\mathbf{l}} \phi_\mathbf{l}(\mathbf{x}),
\end{equation}
with the eigenpairs formulated as:
\begin{equation}
\psi_l(x) = \begin{cases} 
\sqrt{2} \cos(\pi l_1 x_1) & l_2 = 0 \\
\sqrt{2} \cos(\pi l_2 x_2) & l_1 = 0 \\
2 \cos(\pi l_1 x_1) \cos(\pi l_2 x_2) & \text{otherwise}
\end{cases}
\quad , \quad
\lambda_l = (\pi^2 |l|^2 + \tau^2)^{-d}.
\end{equation}
We choose $d=1.2$, $\tau=1.0$ to generate 10k simulations of the Darcy flow problem with a domain size of $128 \times 128$.A histogram of the cell values of the permeability and pressure fields is shown in Figure ~\ref{fig:darcy_hist}.

\subsection{Shallow Water equations}
\begin{figure}[!htb]
    \centering
    \includegraphics[width=\textwidth]{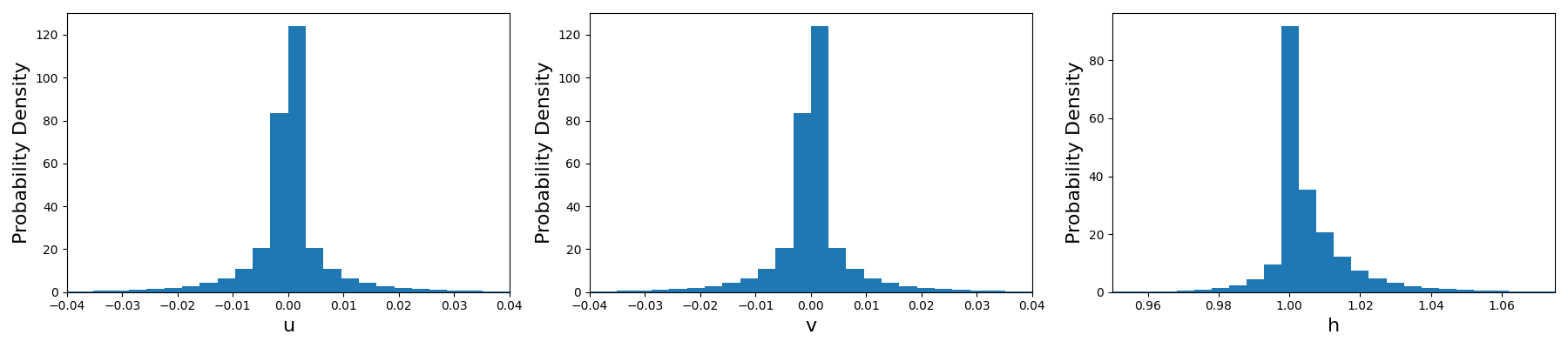}
    \caption{Histogram of the cell values of the velocity fields and the water height of the shallow water equations data.}
    \label{fig:sw_hist}
\end{figure}

We follow the instruction provided in~\cite{cheng2024efficient} to generate 250 simulations, each with 50 snapshots, of the shallow water equations with a domain size of $64 \times 64$. The problem is evolved using the forward in time centered in space (FTCS) scheme. A histogram of the cell values of the velocity fields and the water height is shown in Figure ~\ref{fig:sw_hist}. The $u$ and $v$ components of the water column velocity are initialized at $0.1$ m/s, with a fixed height of $0.1$ mm and a radius of $4$ mm. The spatial domain has dimensions of $50$ mm $\times$ $50$ mm.

\subsection{Diffusion Reaction equations}

\begin{figure}[!htb]
    \centering
    \includegraphics[width=\textwidth]{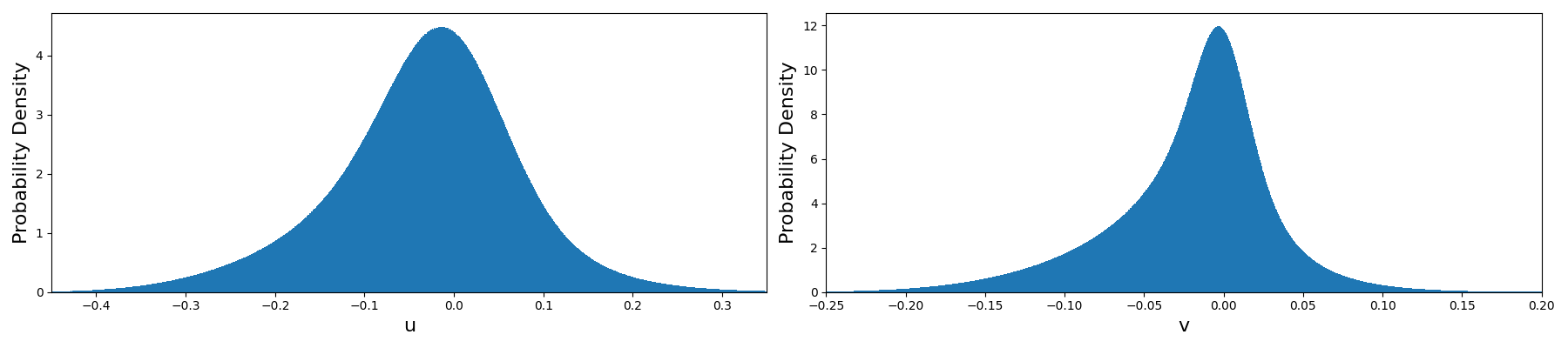}
    \caption{Histogram of the cell values of the activator and inbibitor fields of the diffusion reaction equations data.}
    \label{fig:dr_hist}
\end{figure}

The data is downloaded from the publicly available PDEBench~\cite{takamoto2022pdebench}, The initial concentration profiles of the activator and inhibitor fields are sampled from $\mathcal{N}(0, 1)$. The simulation is performed on a \( 512 \times 512 \) domain with 500 time steps for \( t \in (0, 500] \). The results are downsampled to \( 128 \times 128 \) with 101 timesteps, including the initial condition. A histogram of the cell values of the activator and inhibitor fields is shown in Figure ~\ref{fig:dr_hist}.

\subsection{Compressible Navier-Stokes equations}

\begin{figure}[!htb]
    \centering
    \includegraphics[width=\textwidth]{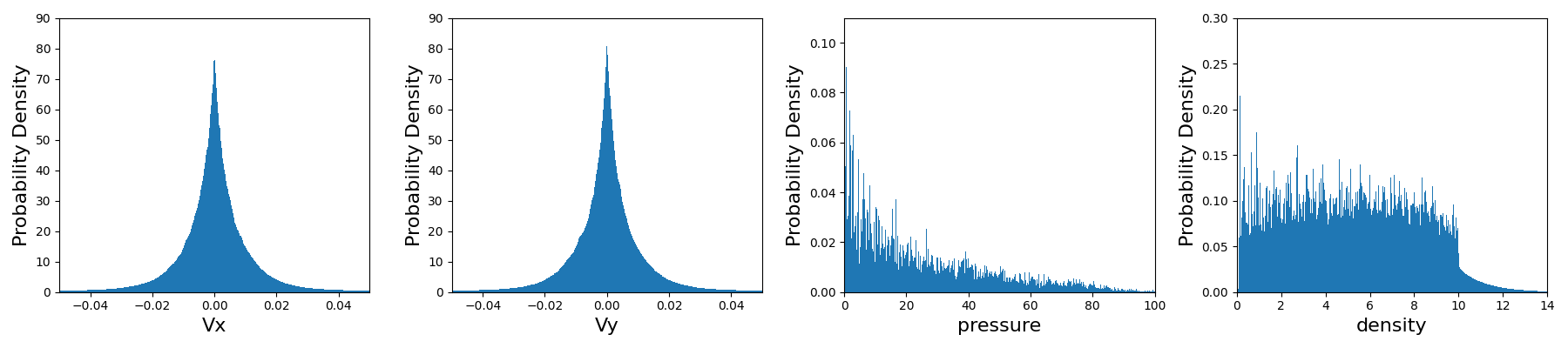}
    \caption{Histogram of the cell values of the four fields of the compressible Navier-Stokes equations training data.}
    \label{fig:ns_hist}
\end{figure}

The data is downloaded from the publicly available PDEBench~\cite{takamoto2022pdebench}. The velocity fields are initialized as follows:
\begin{equation}
\mathbf{v}(x, t=0) = \sum_{i=1}^{n} \bm{A}_i \sin(k_i x + \phi_i),
\label{eq:ns_init}
\end{equation}
where \( n = 4 \), \({|k|} \), and \( k_i = \frac{2\pi n_i}{L} \) are the wave numbers, with \( n_i \) uniformly sampled from \( [1, n_{\text{max}}] \). Here, \( c_s \) is the speed of sound, and \( M \) is the Mach number. The density and pressure are also initialized by adding a uniform background to the perturbation field (Equation~\eqref{eq:ns_init}). The simulation is performed for 20 timesteps with $t \in (0, 2]$. A histogram of the cell values of the four fields is shown in Figure ~\ref{fig:ns_hist}.

\section{Additional Results \label{app:results}}

\subsection{Diffusion model: Multistep sampling \label{app:multistep}}

\begin{table}[!htb]
\centering
\caption{Results of 1000 unseen samples for different PDEs. The diffusion models have an ensemble size of 25 and solved for 20 steps with the predictor-corrector scheme.}
\label{tab:multi_step_validation_results}
\resizebox{\textwidth}{!}{ 
\begin{tabular}{c|c|c|c|c|c}
\toprule
\textbf{PDEs} & \textbf{Obs\%} & \textbf{Metric} & \multicolumn{3}{c}{\textbf{Diffusion Model}} \\ 
\cline{4-6} 
& & & \makecell{\textbf{Guided} \\ \textbf{Sampling}} & \textbf{CFG} & \makecell{\textbf{Cross}\\\textbf{Attention}} \\ 
\midrule
\multirow{9}{*}{\makecell{Shallow \\ water}} 
& \multirow{3}{*}{0.3\%}  & RMSE & \( 7.55 \times 10^{-3} \) & \( 4.81 \times 10^{-3} \) & \(\bm{4.30 \times 10^{-3}}\) \\ 
& & nRMSE & \( \bm{7.66 \times 10^{-1}} \) & \( 1.62 \times 10^{0} \) & \(9.71 \times 10^{-1}\) \\ 
& & cRMSE & \( 8.33 \times 10^{-4} \) & \( 5.07 \times 10^{-4} \) & \(\bm{4.68 \times 10^{-4}}\) \\ 
\cline{2-6}
& \multirow{3}{*}{1\%} & RMSE & \( 7.46 \times 10^{-3} \) & \( 3.72 \times 10^{-3} \) & \(\bm{3.32 \times 10^{-3}}\) \\ 
& & nRMSE & \( 7.45 \times 10^{-1} \) & \( 4.47 \times 10^{-1} \) & \(\bm{3.76 \times 10^{-1}}\) \\ 
& & cRMSE & \( 8.42 \times 10^{-4} \) & \( 3.26 \times 10^{-4} \) & \(\bm{2.99 \times 10^{-4}}\) \\ 
\cline{2-6}
& \multirow{3}{*}{3\%} & RMSE & \( 7.05 \times 10^{-3} \) & \( 3.32 \times 10^{-3} \) & \(\bm{2.85 \times 10^{-3}}\) \\ 
& & nRMSE & \( 6.94 \times 10^{-1} \) & \( 3.35 \times 10^{-1} \) & \(\bm{2.83 \times 10^{-1}}\) \\ 
& & cRMSE & \( 8.52 \times 10^{-4} \) & \( 2.50 \times 10^{-4} \) & \(\bm{2.12 \times 10^{-4}}\) \\ 
\midrule
\multirow{9}{*}{\makecell{Diffusion \\ reaction}} 
& \multirow{3}{*}{0.3\%} & RMSE & \( 7.92 \times 10^{-2} \) & \( 7.61 \times 10^{-2} \) & \(\bm{6.26 \times 10^{-2}}\) \\ 
& & nRMSE & \( 9.67 \times 10^{-1} \) & \( 9.33 \times 10^{-1} \) & \(\bm{7.63 \times 10^{-1}}\) \\ 
& & cRMSE & \( 2.18 \times 10^{-2} \) & \( 1.45 \times 10^{-2} \) & \(\bm{6.07 \times 10^{-3}}\) \\ 
\cline{2-6}
& \multirow{3}{*}{1\%} & RMSE & \( 7.65 \times 10^{-2} \) & \( 7.40 \times 10^{-2} \) & \(\bm{3.98 \times 10^{-2}}\) \\ 
& & nRMSE & \( 9.24 \times 10^{-1} \) & \( 9.00 \times 10^{-1} \) & \(\bm{4.70 \times 10^{-1}}\) \\ 
& & cRMSE & \( 2.07 \times 10^{-2} \) & \( 1.07 \times 10^{-2} \) & \(\bm{4.69 \times 10^{-3}}\) \\ 
\cline{2-6}
& \multirow{3}{*}{3\%} & RMSE & \( 7.12 \times 10^{-2} \) & \( 7.22 \times 10^{-2} \) & \(\bm{2.56 \times 10^{-2}}\) \\ 
& & nRMSE & \( 8.34 \times 10^{-1} \) & \( 8.65 \times 10^{-1} \) & \(\bm{2.59 \times 10^{-1}}\) \\ 
& & cRMSE & \( 1.86 \times 10^{-2} \) & \( 7.90 \times 10^{-3} \) & \(\bm{4.15 \times 10^{-3}}\) \\ 
\midrule
\multirow{9}{*}{\makecell{CFD}} 
& \multirow{3}{*}{0.3\%} & RMSE & \( 2.02 \times 10^{1} \) & \( 6.68 \times 10^{-1} \) & \(\bm{3.55 \times 10^{-1}}\) \\ 
& & nRMSE & \( 9.10 \times 10^{0} \) & \( 8.16 \times 10^{-1} \) & \(\bm{2.43 \times 10^{-1}}\) \\ 
& & cRMSE & \( 2.07 \times 10^{1} \) & \( 4.49 \times 10^{-1} \) & \(\bm{1.56 \times 10^{-1}}\) \\ 
\cline{2-6}
& \multirow{3}{*}{1\%} & RMSE & \( 1.97 \times 10^{1} \) & \( 6.06 \times 10^{-1} \) & \(\bm{2.47 \times 10^{-1}}\) \\ 
& & nRMSE & \( 8.94 \times 10^{0} \) & \( 7.48 \times 10^{-1} \) & \(\bm{1.92 \times 10^{-1}}\) \\ 
& & cRMSE & \( 2.02 \times 10^{1} \) & \( 3.95 \times 10^{-1} \) & \(\bm{1.30 \times 10^{-1}}\) \\ 
\cline{2-6}
& \multirow{3}{*}{3\%} & RMSE & \( 1.83 \times 10^{1} \) & \( 5.02 \times 10^{-1} \) & \(\bm{1.47 \times 10^{-1}}\) \\ 
& & nRMSE & \( 8.47 \times 10^{0} \) & \( 6.03 \times 10^{-1} \) & \(\bm{1.44 \times 10^{-1}}\) \\ 
& & cRMSE & \( 1.87 \times 10^{1} \) & \( 3.17 \times 10^{-1} \) & \(\bm{9.26 \times 10^{-2}}\) \\ 
\midrule
\multirow{6}{*}{Darcy} 
& \multirow{3}{*}{0.3\%} & RMSE & \( 6.72 \times 10^{-1} \) & \( 3.91 \times 10^{-1} \) & \(\bm{2.49 \times 10^{-1}}\) \\ 
& & nRMSE & \( 4.99 \times 10^{-1} \) & \( 2.91 \times 10^{-1} \) & \(\bm{1.86 \times 10^{-1}}\) \\ 
& & cRMSE & \( 9.76 \times 10^{-2} \) & \( 3.54 \times 10^{-2} \) & \(\bm{1.58 \times 10^{-2}}\) \\ 
\cline{2-6}
& \multirow{3}{*}{1.37\%} & RMSE & \( 6.58 \times 10^{-1} \) & \( 3.76 \times 10^{-1} \) & \(\bm{2.01 \times 10^{-1}}\) \\ 
& & nRMSE & \( 4.88 \times 10^{-1} \) & \( 2.80 \times 10^{-1} \) & \(\bm{1.49 \times 10^{-1}}\) \\ 
& & cRMSE & \( 1.04 \times 10^{-1} \) & \( 3.34 \times 10^{-2} \) & \(\bm{1.64 \times 10^{-2}}\) \\ 
\bottomrule
\end{tabular}
}
\end{table}

The results for the diffusion models with the multistep~\cite{lu2022dpm} sampling scheme are shown in Table~\ref{tab:multi_step_validation_results}. These results are generated from an ensemble of 25 trajectories using the predictor-corrector scheme with 20 steps. In general, the cross-attention method demonstrates the best performance in terms of RMSE, nRMSE, and cRMSE across all PDEs.

\subsection{Diffusion model: Generating not Memorizing}

Diffusion models can be prone to memorizing training data. As demonstrated in~\citep{gu2023memorization}, the extent of memorization in image-generation tasks is negatively correlated with the size and diversity of the training data. Additionally,~\citep{carlini2023extracting} showed that in conditional generation tasks, similarities in the conditioning information can exacerbate the problem of memorization in diffusion models. In our case, we design the condition encoding block to also incorporate the position information of the sensing array, which introduces additional variability during training.

To demonstrate that the reconstructed fields are not present in the training data, we perform a t-SNE analysis. We reconstruct fields using $1\%$ of the observed data points from the sensing information in the testing set of the 2D diffusion-reaction equation. These reconstructed fields are then projected using the PCA obtained from the training set, and a t-SNE plot is constructed.

\begin{figure}[!htb]
\centering
\includegraphics[width=\textwidth]{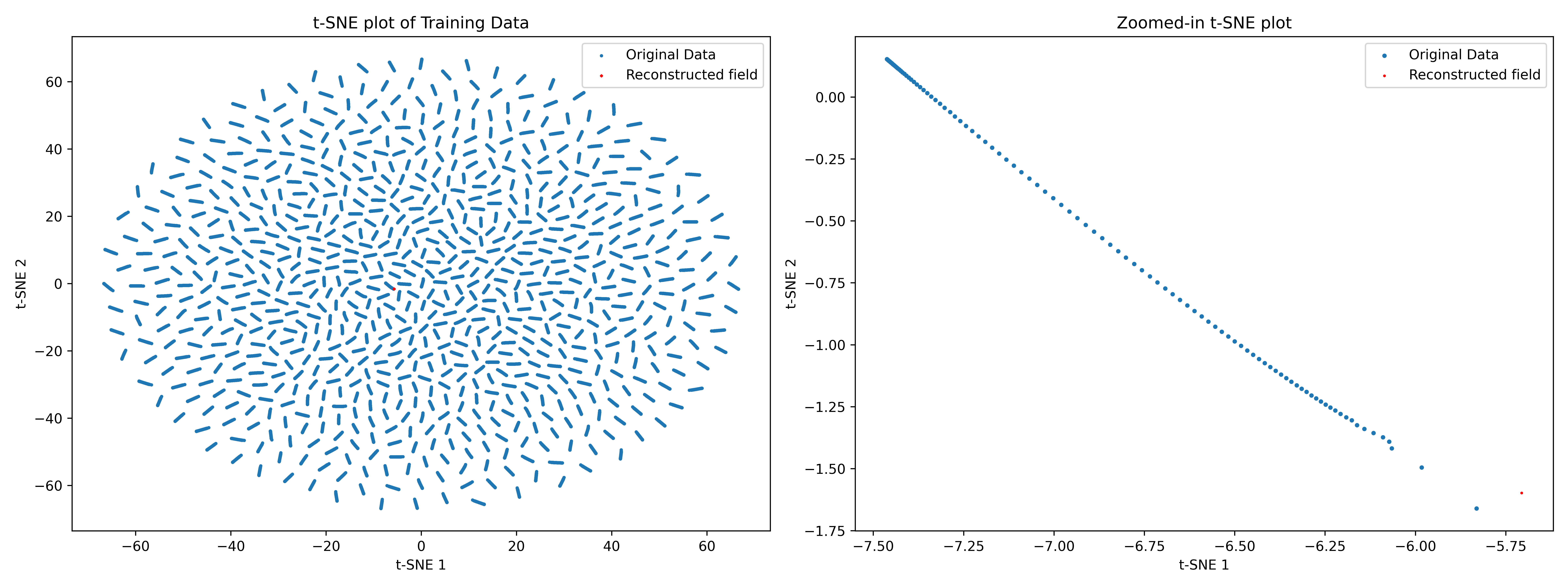}
\caption{t-SNE plot of the PCA space of the diffusion-reaction equation training set. The red dots represent the reconstructed fields from the sensing information, while the blue dots represent the training set.}
\label{fig:tsne}
\end{figure}

\begin{figure}[!htb]
    \centering
    \includegraphics[width=0.7\textwidth]{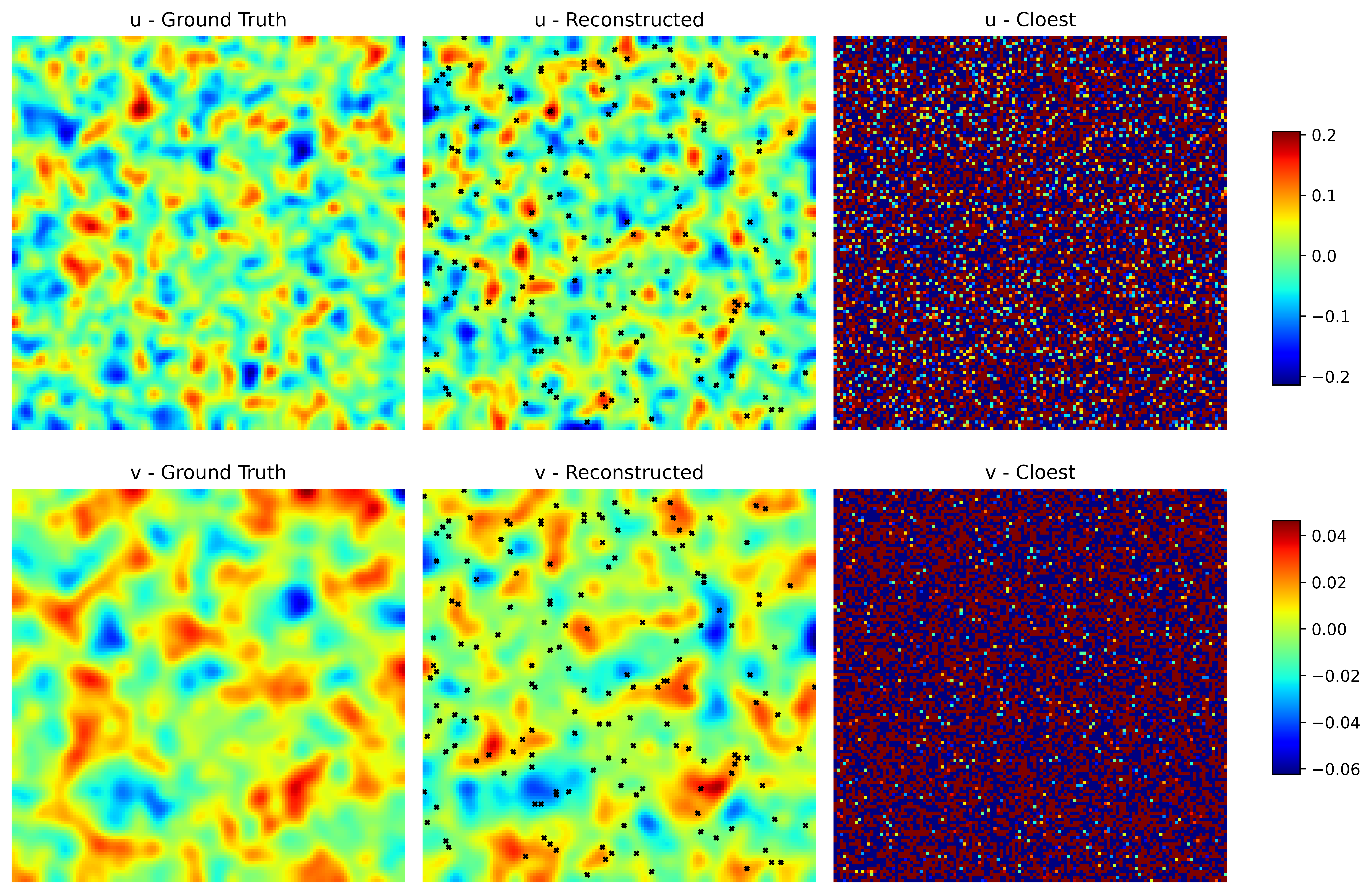}
    \caption{The ground truth, reconstructed fields from the sensing information, and the snapshot that is closest to the reconstructed field in the training set.}
    \label{fig:dr_memory}
\end{figure}

The t-SNE plot of the PCA space of the diffusion-reaction equation training set is shown in Figure ~\ref{fig:tsne}. We select the number of components in the PCA to be 1000, which captures $97\%$ of the variance in the training set. The t-SNE plot reveals that the training dataset is clustered by simulation, with each line-like cluster representing fields from the same simulation.

We measure the $L_2$ distances of the t-SNE representations between the reconstructed fields and the rest of the training data. The ground truth, reconstructed fields, and the closest fields in the training set are shown in Figure ~\ref{fig:dr_memory}. These results indicate that the model does not simply draw from the training data when reconstructing fields but instead generates fields based on the sensing information. However, under the current problem setup, it is challenging to obtain the initial concentration profiles of the reconstructed fields and verify whether they satisfy the standard Gaussian initialization.

\subsection{0.3\% observed points with noisy observations}

\begin{figure}[!htb]
    \centering
    \includegraphics[width=\textwidth]{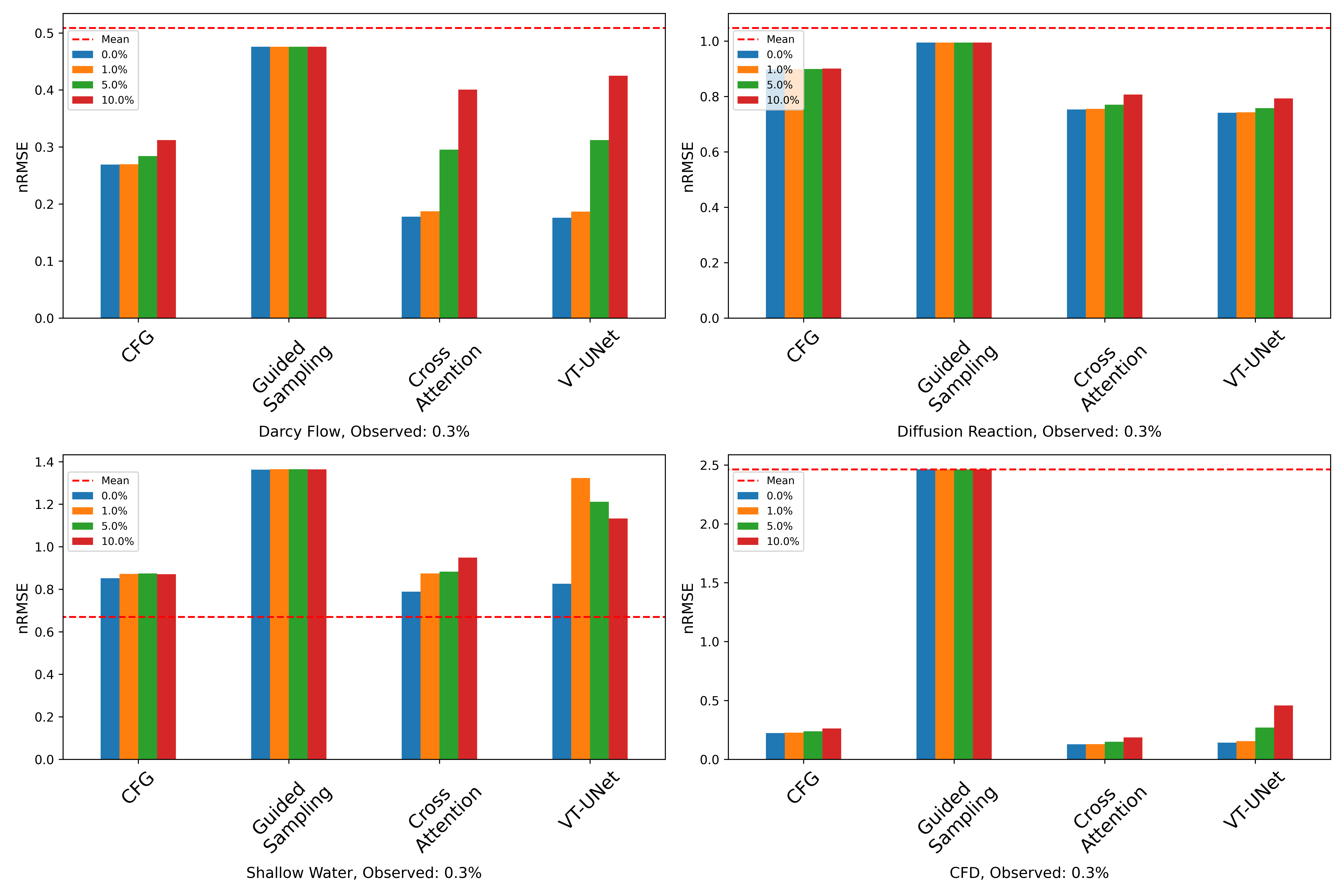}
    \caption{Bar plot of nRMSE for the PDEs with 0.3\% observed data points and various observation noise levels. The red dashed line denotes the error of reconstructing the field with the mean of the training data. The diffusion models are configured with 20 steps, with a predictor-corrector scheme and an ensemble of 25 trajectories.}
    \label{fig:bar_chart_0.03}
\end{figure}

The extreme case of sparse observations, with 0.3\% observed data and various observation noise levels, is shown in Figure ~\ref{fig:bar_chart_0.03}. Under these circumstances, the diffusion models with cross-attention typically perform slightly better than the VT-UNet. Additionally, the diffusion models with CFG can occasionally outperform the cross-attention method, possibly because the learned encoded condition is more generalized than that of the cross-attention method. However, all methods struggle to reconstruct the fields of the shallow water equations. This difficulty arises because the wavelet patterns in the fields leave a majority of the domain blank, and under such extreme sparsity, the uniformly sampled observations are less likely to fall on the wavelet patterns.

\subsection{Varying resolution\label{app:res}}

CNN-based models have been shown to generalize well across different resolutions~\cite{fukami2024data, yasuda2023rotationally}, and the backbone of our diffusion model is UNet, which is an variant of CNN with skip connection at different hierarchical levels. However, in our investigation with the higher resolution darcy flow problem, we found that both VT-UNet and the conditional diffusion model fails to generalize. This is likely due to the changes in feature scale perceived by the convolutional kernels as the resolution increases. To address this issue, popular approaches include scaling input, training with multiple resolutions (data augmentation), and fine-tuning the model on the target resolution. 

\begin{figure}[!htb]
    \centering
    \includegraphics[width=0.8\linewidth]{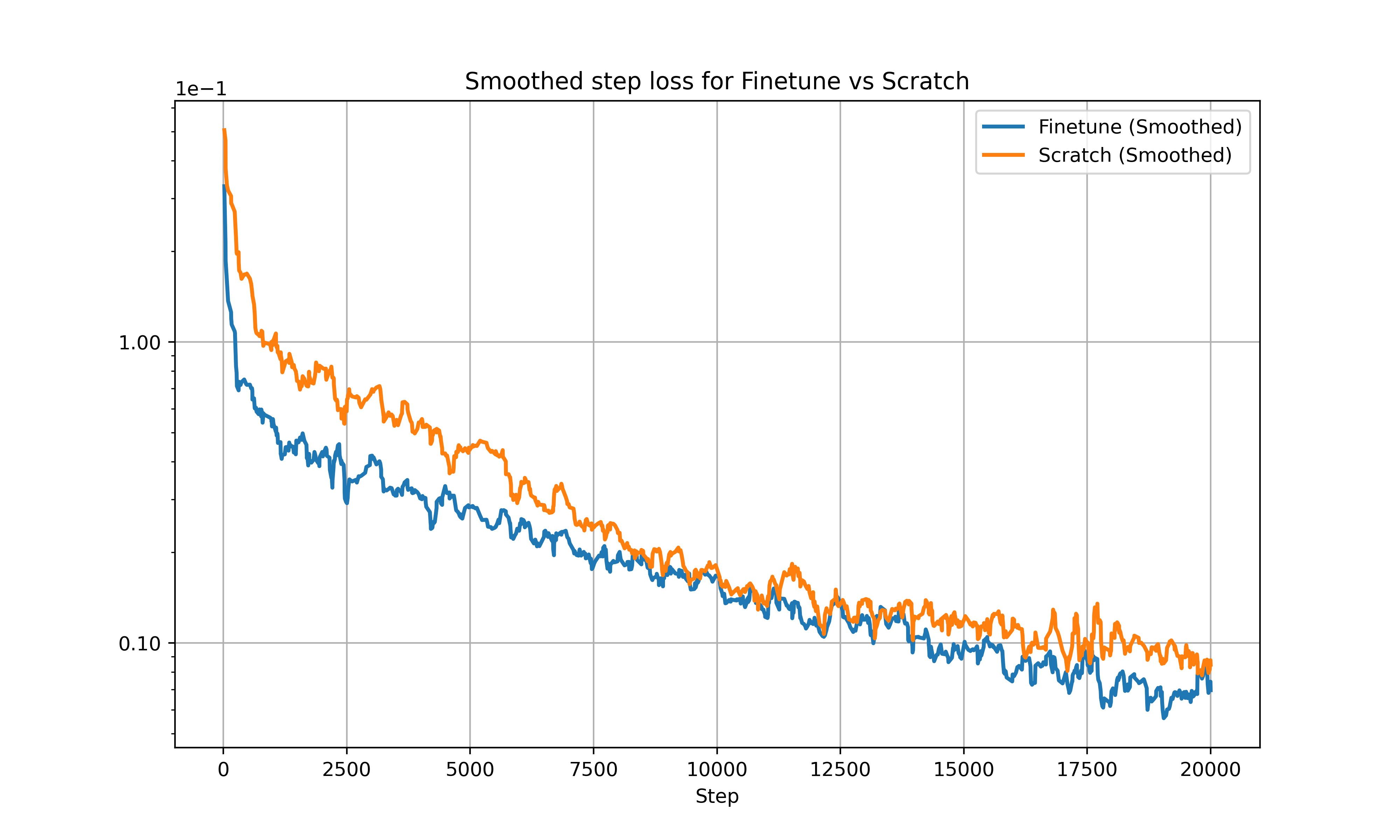}
    \caption{Step loss comparison between training from scratch and fine-tuning on the Darcy flow problem with resolution of $256 \times 256$. The fine-tuning process shows a faster convergence rate. Step loss is smoothed with a moving average of 10 steps.}
    \label{fig:scatch_vs_finetune}
\end{figure}

\begin{table}[!htb]
\centering
\caption{Comparison of training from scratch and fine-tuning on the Darcy flow problem with resolution of $256 \times 256$ on the 1k test data with an ensemble size of 25. The number of sampled points is 256, which is equivalent to the 1.37\% in $128 \times 128$ resolution.}
\label{tab:darcy_comparison_finetune}
\begin{tabular}{c c c c c}
\toprule
 Model & Sampling & RMSE & nRMSE & cRMSE \\ 
\midrule
Scratch & Predictor-Corrector & 0.177  & 0.133 & 0.018  \\ 
Scratch & Multistep & 0.321  & 0.234 & 0.075  \\ 
\midrule
Finetune &  Predictor-Corrector & 0.249 & 0.187 & 0.019 \\ 
Finetune &  Multistep & $\bm{0.166}$ & $\bm{0.123}$ & $\bm{0.016}$ \\ 
\bottomrule
\end{tabular}
\end{table}

The $256 \times 256$ resolution Darcy flow problem is generated with the same hyperparameters and seeding as the $128 \times 128$ resolution. Both models are trained for 20,000 steps with a learning rate of $1e-5$. The smoothed step loss is shown in Figure~\ref{fig:scatch_vs_finetune}. The fine-tuning process demonstrates a faster convergence rate. We compare the performance of training from scratch and fine-tuning from the $128 \times 128$ resolution in Table~\ref{tab:darcy_comparison_finetune}. Interestingly, the model trained from scratch shows better performance with the predictor-corrector sampling method, which aligns with the findings in the main text. However, the trend reverses for the fine-tuned model, where the multistep method outperforms the predictor-corrector method. This indicates that the trained model can be adapted to different resolutions, and sampling methods should be revised accordingly.

\subsection{Colored noise \label{app:color_noise}}

\begin{figure}[!htb]
    \centering
    \includegraphics[width=0.8\linewidth]{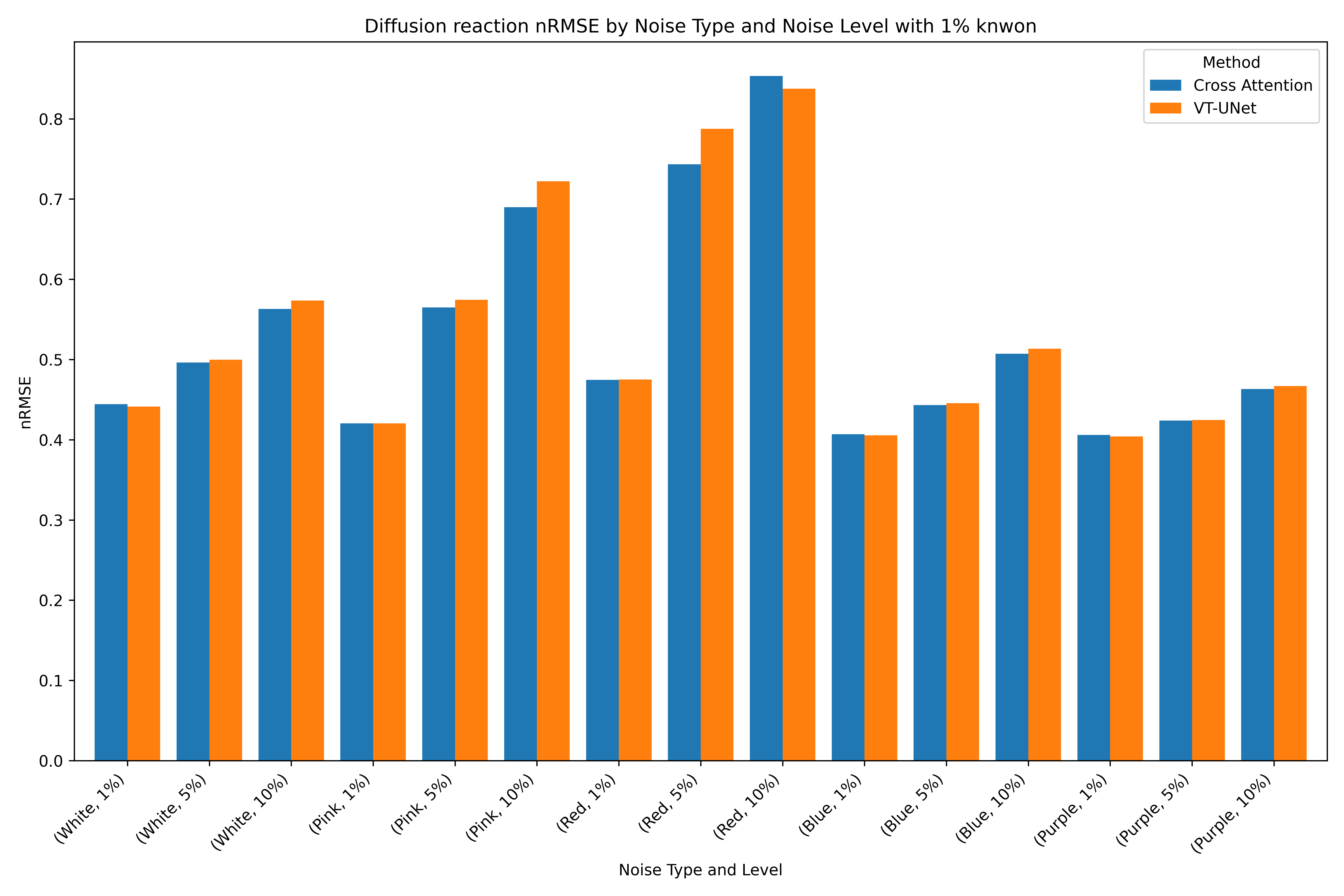}
    \caption{Colored noise performance comparison for the diffusion reaction problem. Results of nRMSE are computed with 1k test data with an ensemble size of 25 and 1\% known values.}
    \label{fig:color_noise}
\end{figure}

Results of 1D diffusion-reaction problems with colored noise are shown in Figure~\ref{fig:color_noise}. The performance of the diffusion model is evaluated with respect to noise levels, where the nRMSE is computed for 1k test data with an ensemble size of 25 and 1\% known values. In general, the same tendency is observed across different types of noise, where the performance of the diffusion model surpasses the VT-UNet baseline as the noise level increases.

\subsection{Shallow water energy plot\label{app:energy}}

\begin{figure}[!htb]
    \centering
    \includegraphics[width=0.8\linewidth]{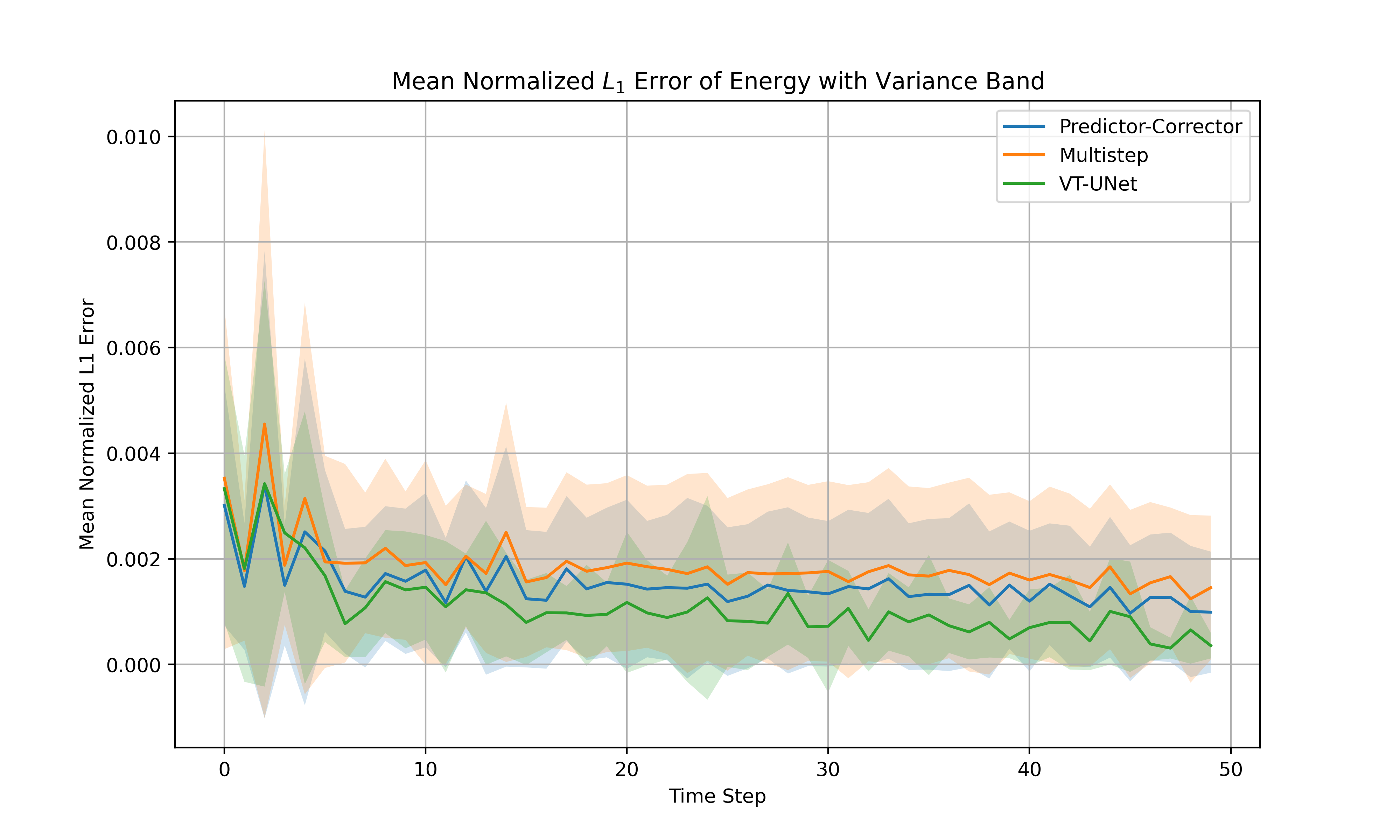}
    \caption{Energy plot of shallow water equations on 10 test set trajectories, with 1\% values known, the cross-attention diffusion model is utilized with an ensemble size of 25 and 20 reverse sampling steps.}
    \label{fig:energy_plot}
\end{figure}

In Figure~\ref{fig:energy_plot}, we present the normalized energy plot for the shallow water equations on 10 test set trajectories, with 1\% known values, 0\% noise, and the cross-attention diffusion model employing an ensemble of 25 and 20 reverse sampling steps. This plot supports the findings in the main text, where nRMSE results show that the VT-UNet model performs better under noise-free conditions. The low RMSE observed in these noise-free scenarios aligns with this energy-based evaluation, highlighting the model’s effectiveness in capturing fine-scale features, even with sparse observations.
\end{document}